\documentclass{article}
\usepackage{iclr2027_conference,times}

\usepackage{amsmath,amsfonts,bm}

\def\eqref#1{equation~\ref{#1}}

\def\1{\bm{1}}

\def\rb{{\textnormal{b}}}

\DeclareMathAlphabet{\mathsfit}{\encodingdefault}{\sfdefault}{m}{sl}
\SetMathAlphabet{\mathsfit}{bold}{\encodingdefault}{\sfdefault}{bx}{n}

\newcommand{\E}{\mathbb{E}}

\newcommand{\R}{\mathbb{R}}

\DeclareMathOperator*{\argmax}{arg\,max}
\DeclareMathOperator*{\argmin}{arg\,min}

\usepackage{hyperref}
\usepackage{url}

\usepackage[legacy]{mycommands}
\usepackage{threeparttable}
\usepackage{multirow}
\usepackage{booktabs}
\usepackage{makecell}
\usepackage{paralist}
\usepackage{algorithmic}
\usepackage{algorithm}
\usepackage{subfig}
\usepackage{graphicx}
\usepackage{wrapfig}

\title{Communication-Efficient Agnostic Federated Learning via Faster Convergence and \\ Compression}

\author{
Haomin Bai$^{1,2}$ \quad
Junyan Sun$^{1,2}$ \quad
Sifan Yang$^{1,2}$ \quad
Bo Xue$^3$ \quad
Lijun Zhang$^{1,2}$\thanks{Lijun Zhang is the corresponding author.}\\
$^1$State Key Laboratory of Novel Software Technology, Nanjing University, Nanjing, China\\
$^2$School of Artificial Intelligence, Nanjing University, Nanjing, China\\
$^3$Department of Computer Science, City University of Hong Kong, Hong Kong, China\\
\texttt{\{baihm,sunjy,yangsf,zhanglj\}@lamda.nju.edu.cn}\\
\texttt{boxue4-c@my.cityu.edu.hk}
}

\iclrfinalcopy
\begin{document}

\maketitle

\begin{abstract}
Agnostic federated learning (AFL) seeks a model that performs reliably across $m$ heterogeneous workers, but communication remains a bottleneck. 
We improve communication efficiency by reducing the number of synchronization rounds via faster convergence and the communication cost per round via compression.
We first propose \textsc{AFL-BR}, which updates the dual weights over workers using online mirror ascent with KL divergence and blockwise restarts.
It achieves an $O((\log m)^{1/4}T^{-1/8})$ stationarity rate after $T$ update rounds, reducing the $m$-dependence of the synchronization rounds required for convergence from polynomial to logarithmic order.
Building on \textsc{AFL-BR}, we develop \textsc{AFL-Com} by applying bidirectional compression with error feedback (EF).
Instead of compressing local gradients, workers apply EF to their dual-weighted gradients, enabling direct control of the aggregated compression error under time-varying weights.
We then establish an $O((\delta^{-1}+(\log m)^{1/4})T^{-1/8})$ stationarity rate for \textsc{AFL-Com} under general $\delta$-approximate compressors and improve the $\delta$-dependence from $\delta^{-1}$ to $\delta^{-1/2}$ for additive-and-idempotent compressors with shared randomness (SR).
With suitable compression levels, \textsc{AFL-Com} retains the same convergence rate as \textsc{AFL-BR} at a lower per-round communication cost, yielding reductions in total communication complexity by factors of $(\log m)^{1/4}$ with Top-$k$ and $(\log m)^{1/2}$ with Rand-$k$ and SR.
Experiments validate the improved synchronization and communication efficiency of our methods.
\end{abstract}

\section{Introduction}\label{sec:introduction}
Federated learning (FL) enables multiple workers to collaboratively train a model while keeping data distributed~\citep{2017_FL_McMahan}, making it attractive for privacy-sensitive applications and large-scale model training~\citep{2020_FLreview_app}. Classical FL methods learn a global model by minimizing the average loss over $m$ workers. In practice, however, data distributions across workers can be highly heterogeneous~\citep{2023_survey_HFL}, making an average-optimal model vulnerable to distribution shifts and unfair across workers~\citep{2019AFL}. These limitations are particularly concerning in high-stakes applications such as healthcare~\citep{2023_Nature_Fairness} and finance~\citep{2020_OpenBank}.

To address these issues,~\citet{2019AFL} propose agnostic FL (AFL), which seeks a model that performs robustly across arbitrary mixtures of local data distributions.
Let $\Delta_m=\{\q\in\R^m:\q\geq\mathbf 0_m,\mathbf 1_m^\top\q=1\}$ denote the probability simplex. AFL solves the minimax problem
\begin{equation} \label{eq:gdro}
    \min_{\w \in \W} \max_{\q \in \Delta_m}  \   \left\{\phi(\w,\q)= \sum_{i=1}^m q_i R_i(\w) \right\},
\end{equation}
where $\q$ denotes the dual weights over $m$ workers and $R_i(\w) = \E_{\z \sim \P_i}[\ell(\w;\z)]$ is the local objective of worker $i$, namely, the expected loss of model $\w\in\R^d$ on samples drawn from the local data distribution $\P_i$. 
The same objective also appears in group distributionally robust optimization (GDRO), with workers interpreted as groups~\citep{Gouop_DRO_ICLR_2020}. We focus on smooth and possibly nonconvex local objectives over $\W=\R^d$, as commonly encountered in neural-network training.

Despite the robustness benefits of AFL, communication overhead remains a major bottleneck to its scalability and efficiency~\citep{2021_Survey_FL}. 
We measure communication complexity by the total amount of information exchanged between workers and the server to attain an $\varepsilon$-stationary solution.
It is determined by the number of synchronization rounds and the communication cost per round, which can be reduced through faster convergence and compression, respectively.
Existing nonconvex AFL methods, however, remain limited in both respects:
\begin{compactenum}
    \item First, existing guarantees exhibit polynomial dependence on $m$~\citep{2020_GDA_minimax,2020_DRFA}, causing the number of synchronization rounds to grow rapidly with the federation size. Although~\citet{2020_DRFA} reduces synchronization frequency via local updates, its slower convergence requires more rounds than its fully synchronized specialization.
    \item Second, existing work on compressed AFL only supports uplink communication and is restricted to unbiased compressors~\citep{2026_unlocking_ICLR}. Moreover, its nonconvex guarantee relies on a Minty-inspired global condition, which is generally difficult to verify. Bidirectional compression with general contractive compressors remains unexplored.
\end{compactenum} 

In this work, we address both sources of communication overhead in turn. 
First, we propose \textsc{AFL-BR}, a blockwise-restarted algorithm that updates the dual weights using online mirror ascent with KL divergence~\citep{2016_Hazan_book}.
The main challenge in analysis is to control a dual-gap term with a time-varying comparator while exploiting the KL geometry, whose $\ell_1/\ell_\infty$ structure avoids the polynomial dimension factors of Euclidean analysis.
Since the optimal dual weights vary with the model iterate, static-regret guarantees for a fixed comparator do not apply directly.
To handle the changing comparators, we partition the iterations into blocks and use the block-start maximizer as a fixed comparator, decomposing the dual gap into a comparator-mismatch term controlled by model movement and a static-regret term.
A further challenge is that the resulting static-regret bound depends on the block-initial KL divergence, which can be arbitrarily large.
We therefore restart the dual weights from the uniform distribution at each block boundary, ensuring the initial KL divergence is bounded by $\ln m$.
Together, we establish an $O((\log m)^{1/4}T^{-1/8})$ stationarity guarantee, which yields $O((\log m)^2\varepsilon^{-8})$ synchronization rounds for finding an $\varepsilon$-stationary solution, reducing their dependence on $m$ from at least quadratic~\citep{2020_GDA_minimax,2020_DRFA} to logarithmic.

\begin{table*}[t]
\centering
\caption{\textbf{Comparison of nonconvex AFL methods using Top-$k$ and Rand-$k$ as examples.}
$d$ is the model dimension and $m$ the number of workers. Definitions of compressors are provided in Appendix~\ref{app:exam_compressors}.
$\star,\clubsuit$ Hidden $m$-dependence is made explicit; see Appendices~\ref{app:sgda_hidden_m} and~\ref{app:drfa_refined_gamma}. 
$\S$ Set $k=\Theta(d(\log m)^{-1/4})$ in Theorem~\ref{thm:aflcom_br_general}.
$\ast$ Set $k=\Theta(d(\log m)^{-1/2})$ with SR in Theorem~\ref{thm:aflcom_br_ai}. 
\citet{2026_unlocking_ICLR} is omitted since it relies on a Minty-inspired condition with a different criterion.}
\label{tab:AFL-summary}
\scriptsize
\begin{tabular}{ccccc}
\toprule
\textbf{Algorithm} & \textbf{Technique} & \makecell{\textbf{Synchronization rounds}\\\textbf{for $\varepsilon$-stationarity}} & \makecell{\textbf{Communication cost}\\\textbf{per round}} & \makecell{\textbf{Communication}\\\textbf{complexity}} \\
\midrule
\textsc{SGDA}$^{\star}$ & -- & $O(m^{7/2}\varepsilon^{-8})$ & $O(dm)$ & $O(dm^{9/2}\varepsilon^{-8})$ \\
\midrule
\multirow{2}{*}{\textsc{DRFA}$^{\clubsuit}$}
& Local updates & $O(\varepsilon^{-12}+m^2\varepsilon^{-8})$ & $O(dm)$ & $O(dm\varepsilon^{-12}+dm^3\varepsilon^{-8})$ \\
\cmidrule(lr){2-5}
& Fully synchronized & $O(m^2\varepsilon^{-8})$ & $O(dm)$ & $O(dm^3\varepsilon^{-8})$ \\
\midrule
\makecell{\textsc{AFL-BR}\\(Algorithm~\ref{alg:uafl_br})} & -- & $O((\log m)^2\varepsilon^{-8})$ & $O(dm)$ & $O(dm(\log m)^2\varepsilon^{-8})$ \\
\midrule
\multirow{2}{*}{\makecell{\textsc{AFL-Com}\\(Algorithm~\ref{alg:aflcom_br})}} & \makecell{Compression\\(Top-$k$)$^{\S}$} & $O((\log m)^2\varepsilon^{-8})$ & $O(dm(\log m)^{-1/4})$ & $O(dm(\log m)^{7/4}\varepsilon^{-8})$ \\
\cmidrule(lr){2-5}
& \makecell{Compression\\(Rand-$k$ with SR)$^{\ast}$} & $O((\log m)^2\varepsilon^{-8})$ & $O(dm(\log m)^{-1/2})$ & $O(dm(\log m)^{3/2}\varepsilon^{-8})$ \\
\bottomrule
\end{tabular}
\end{table*}

For the second objective, the dominant per-round communication cost arises from transmitting high-dimensional gradients used to update the model, whereas the dual update exchanges only $O(m)$ scalar entries.
Building on \textsc{AFL-BR}, we develop \textsc{AFL-Com}, which applies \emph{bidirectional} compression with error feedback (EF)~\citep{EF_ICML19} to the vectors exchanged for model updates.
Since AFL aggregates local gradients using time-varying dual weights, standard EF on local gradients cannot directly control the global compression error. \textsc{AFL-Com} therefore applies EF to dual-weighted local gradients, enabling sharp control of the aggregated error.
Under general $\delta$-approximate compressors, we establish an $O((\delta^{-1}+(\log m)^{1/4})T^{-1/8})$ stationarity rate. 
Choosing $\delta=\Omega((\log m)^{-1/4})$ keeps the compression error no larger than the term already present in \textsc{AFL-BR}, preserving the order of synchronization rounds for convergence while reducing the communication cost per round.
For Top-$k$ sparsification~\citep{2018_SparSGD_toprandk}, which retains the $k$ coordinates with largest absolute values, choosing $k=\Theta(d(\log m)^{-1/4})$ reduces the per-round cost from $O(dm)$ to $O(dm(\log m)^{-1/4})$ and the communication complexity from $O(dm(\log m)^2\varepsilon^{-8})$ of \textsc{AFL-BR} to $O(dm(\log m)^{7/4}\varepsilon^{-8})$.

We further show that, when using additive-and-idempotent compressors with shared randomness (SR), the downlink compression error vanishes, yielding a sharper $O((\delta^{-1/2}+(\log m)^{1/4})T^{-1/8})$ stationarity rate. 
The corresponding squared-norm bound has the same $\delta^{-1}$ dependence as the lower-bound term for fixed-weight distributed nonconvex optimization~\citep{FCC-NIPS22}.
For Rand-$k$ sparsification~\citep{2018_SparSGD_toprandk} with SR, setting $k=\Theta(d(\log m)^{-1/2})$ reduces the communication complexity to $O(dm(\log m)^{3/2}\varepsilon^{-8})$, improving over \textsc{AFL-BR} by a factor of $(\log m)^{1/2}$. Table~\ref{tab:AFL-summary} summarizes these guarantees. 
Our contributions are summarized as follows:
\begin{compactenum}
\item We propose \textsc{AFL-BR}, which updates the dual weights using KL-based online mirror ascent with blockwise restarts. It achieves a stationarity rate of $O((\log m)^{1/4}T^{-1/8})$, reducing the $m$-dependence of synchronization rounds from polynomial to logarithmic order. The same approach also yields a new algorithm and convergence analysis for nonconvex GDRO.
\item Building on \textsc{AFL-BR}, we develop \textsc{AFL-Com}, which uses bidirectional compression with EF. Under general $\delta$-approximate compressors, it achieves an $O((\delta^{-1}+(\log m)^{1/4})T^{-1/8})$ stationarity rate. With Top-$k$ and $k=\Theta(d(\log m)^{-1/4})$, it reduces the communication complexity by a factor of $(\log m)^{1/4}$ relative to \textsc{AFL-BR}.
\item For additive-and-idempotent compressors with SR, we show that the downlink compression error vanishes, improving the $\delta$-dependence of the stationarity rate from $\delta^{-1}$ to $\delta^{-1/2}$. With Rand-$k$ using $k=\Theta(d(\log m)^{-1/2})$ and SR, \textsc{AFL-Com} further reduces the communication complexity by a factor of $(\log m)^{1/2}$ relative to \textsc{AFL-BR}.
\item Experiments demonstrate the synchronization efficiency of \textsc{AFL-BR} and show that \textsc{AFL-Com} further reduces the total communication cost.
\end{compactenum}

\section{Related Work} \label{sec:related}
\subsection{Agnostic Federated Learning and GDRO}
AFL can be viewed as a worker-level instance of GDRO, with workers serving as predefined groups~\citep{2019AFL,Gouop_DRO_ICLR_2020}. In the centralized setting, relevant GDRO methods include stochastic mirror descent~\citep{nemirovski-2008-robust,Carmom_NIPS22_clipSMD} and two-player approaches based on online learning~\citep{2022_Soma_DRO,Online:Multiple:Distribution,zhang2023-SA-GDRO,Bai2025FSQ,2026_GDRO_TPAMI_Zhang}. However, their theoretical guarantees rely on convex losses and do not directly extend to the smooth nonconvex AFL setting considered here.

Nonconvex AFL can be addressed through general nonconvex--concave minimax methods such as SGDA~\citep{2020_GDA_minimax}, which achieves an $O(T^{-1/8})$ stationarity rate. To reduce communication, \citet{2020_DRFA} propose \textsc{DRFA}, which performs multiple local updates between periodic dual updates. While this design reduces the number of synchronization rounds to $O(T^{3/4})$, it slows the stationarity rate to $O(T^{-1/16})$, resulting in even more synchronization rounds to reach a target accuracy.
Faster rates are possible under stronger structures or assumptions, including $O(T^{-1/6})$ using nested inner loops~\citep{2022_Rafique_minmax_doubleloop} and $O(T^{-1/4})$ under strict-complementarity and bounded-iterate assumptions~\citep{2024_General_minmax_FESSGDA}.
More recently, \citet{2026_unlocking_ICLR} incorporate communication compression into AFL, but their nonconvex guarantee relies on a Minty-inspired global condition and restricts compression to the worker-to-server direction and unbiased compressors.
Additional results for convex AFL and variants are deferred to Appendix~\ref{app:related_work}.

\subsection{Communication Compression}
Communication compression methods can be broadly categorized into quantization~\citep{2022_quantization_NC} and sparsification~\citep{2018_sparsification_1}. 
They have been widely used in distributed optimization~\citep{2017_QSGD,2022_EF_FL_1,2022_EF_FL_2} and federated learning~\citep{2021_FedCOM}.
A key mechanism for convergence under compression is error feedback (EF)~\citep{seide14_interspeech,EF_ICML19}, which maintains an accumulated compression error and adds it back to subsequent updates. 
EF21~\citep{2021_EF21,2023_EF21P_Fri} is a variant of EF that tracks the transmitted information and compresses the resulting residual.
In the parameter-server setting, compression has been applied to the uplink (from workers to the server)~\citep{2020_FedPAQ,2020_QLSGD,2021_EF21,2026_unlocking_ICLR}, the downlink (from the server to workers)~\citep{2024_MARINAP}, and in both directions~\citep{2020_RCEFL_both,2021_WirelessFL_Bid,2022_NIPS_MASHA,2023_EF21P_Fri}. 
The most closely related work is~\citet{2026_unlocking_ICLR}, which studies compressed AFL but only for uplink communication with unbiased compressors.
Moreover, \citet{2023_GDRO_dec_cp} study compressed AFL under dual regularization in a decentralized setting, which differs from the parameter-server setting considered here.

Regarding fundamental limits, smooth nonconvex distributed stochastic optimization with contractive compression admits a lower bound of $\Omega(\delta^{-1}T^{-1}+m^{-1/2}T^{-1/2})$ on the expected squared gradient norm~\citep{FCC-NIPS22}. 
Fixing $\q=m^{-1}\mathbf 1_m$ reduces AFL to fixed-weight distributed optimization, so this lower bound provides a benchmark for the dependence on $\delta$.

\section{Fast Convergence for Nonconvex AFL}\label{sec:uafl}
In this section, we study AFL with smooth nonconvex local objectives and develop \textsc{AFL-BR}, which reduces the $m$-dependence of the stationarity guarantee from polynomial to logarithmic order.

\subsection{Problem Setup and Existing Guarantees}\label{sec:nonconvex_setup}
For problem~(\ref{eq:gdro}), define the worst-case objective $\Phi(\w)=\max_{\q\in\Delta_m}\phi(\w,\q)$, where $\phi(\w,\q)=\sum_{i=1}^m q_iR_i(\w)$.
Let $\mathcal F_t$ contain all randomness revealed before the stochastic oracle queries at round $t$, and define $\mathbb E_t[\cdot]=\mathbb E[\cdot\mid\mathcal F_t]$.
Since $\Phi$ is generally nonsmooth even when each $R_i(\cdot)$ is smooth, we measure convergence through the gradient of its Moreau envelope~\citep{2019_Moreau}, a standard criterion in nonconvex--concave minimax optimization~\citep{2020_GDA_minimax,2020_DRFA}.
\begin{definition}[Moreau envelope]\label{def:moreau_envelope} 
For a function $\Phi:\R^d\to\R$ and a parameter $\lambda>0$, its Moreau envelope is defined as $\Phi_\lambda(\x)=\min_{\y\in\R^d}\left\{\Phi(\y)+\frac{1}{2\lambda}\|\y-\x\|_2^2\right\}$ for $\x\in\R^d$. 
\end{definition}
We impose the following assumptions for the nonconvex analysis.
\begin{assumption}\label{ass:nonconvex_smooth}
For every $i\in[m]$, the local objective $R_i$ is differentiable and $L$-smooth, i.e., $\|\nabla R_i(\w)-\nabla R_i(\w')\|_2\leq L\|\w-\w'\|_2$ for all $\w,\w'\in\R^d$.
\end{assumption}
\begin{assumption}\label{ass:nonconvex_lower}
The function $\Phi(\w)$ is lower bounded, i.e., $\Phi_*=\inf_{\w\in\R^d}\Phi(\w)>-\infty$.
\end{assumption}
An output \(\w\) is called an $\varepsilon$-stationary solution if $\mathbb{E}[\|\nabla\Phi_{1/(2L)}(\w)\|_2]\leq\varepsilon$.
We further impose the following assumptions on the stochastic oracles used for updates.
\begin{assumption}\label{ass:nonconvex_primal_oracle}
For each $i\in[m]$ and any $\w\in\mathbb R^d$, a sample $\z\sim\P_i$ produces $\nabla\ell(\w;\z)$ satisfying
\begin{equation}\label{eqn:nonconvex_primal_oracle}
\E_{\z\sim\P_i}[\nabla\ell(\w;\z)]=\nabla R_i(\w),\ \|\nabla R_i(\w)\|_2\leq G,\ \E_{\z\sim\P_i}\left[\|\nabla\ell(\w;\z)-\nabla R_i(\w)\|_2^2\right]\leq\sigma_w^2.
\end{equation}
For convenience, we denote $\Gamma_w^2:=G^2+\sigma_w^2$, which bounds the stochastic-gradient second moment.
\end{assumption}
\begin{assumption}\label{ass:nonconvex_dual_oracle}
For every $i\in[m]$ and $\w\in\mathbb R^d$, let $\xi_i(\w;\z)=\ell(\w;\z)-R_i(\w)$. There exists $\sigma_q>0$ such that
$\E_{\z\sim\P_i}[\xi_i(\w;\z)]=0$ and $\E_{\z\sim\P_i}\left[\exp\left(\lambda\xi_i(\w;\z)\right)\right]\leq\exp(\lambda^2\sigma_q^2/2)$ for all $\lambda\in\mathbb R$.
\end{assumption} 
Assumption~\ref{ass:nonconvex_dual_oracle} constrains only the centered stochastic-loss fluctuation and does not require a uniform bound on either $R_i(\w)$ or $\ell(\w;\z)$, in contrast to the uniform boundedness assumptions used in prior AFL methods~\citep{2019AFL,2020_DRFA}.
Moreover, Assumption~\ref{ass:nonconvex_dual_oracle} is satisfied by Gaussian noise with variance at most $\sigma_q^2$ and by any centered noise supported on an interval of width at most $2\sigma_q$ via Hoeffding's lemma~\citep{cesa2006prediction}.

\textbf{Existing guarantees and motivation.}
Existing guarantees are stated in terms of global Euclidean quantities whose dependence on $m$ is implicit, potentially obscuring substantial scalability costs.
Accounting for this dependence reveals polynomial scaling in $m$. In particular, \textsc{SGDA}~\citep{2020_GDA_minimax} specialized to AFL yields an $O(m^{7/16}T^{-1/8})$ stationarity rate, while fully synchronized \textsc{DRFA}~\citep{2020_DRFA} improves the worker dependence to $O(m^{1/4}T^{-1/8})$, corresponding to $O(m^2\varepsilon^{-8})$ synchronization rounds.
Moreover, its local-update variant has a synchronization-round bound of $O(\varepsilon^{-12}+m^2\varepsilon^{-8})$.
These polynomial $m$-factors arise from joint smoothness and oracle variance in \textsc{SGDA} and dual-gradient bounds in \textsc{DRFA}, motivating an analysis that avoids such $m$-dependent Euclidean quantities.
See Appendix~\ref{app:table_details} for details.

\subsection{\textsc{AFL-BR}: A Blockwise-Restarted Algorithm for Nonconvex AFL}\label{sec:nonconvex_gdro}
The $m$-dependence of the convergence guarantee is governed by the cumulative dual gap arising in the Moreau-envelope analysis, namely $\sum_{t=1}^T\varepsilon_t^q$, where $\varepsilon_t^q=\phi(\w_t,\q_t^*)-\phi(\w_t,\q_t)$ for $\q_t^*\in\argmax_{\q\in\Delta_m}\phi(\w_t,\q)$.
If the comparator $\q_t^*$ is fixed, the term $\sum_{t=1}^T\varepsilon_t^q$ would reduce to static regret, which can be controlled by KL-based online mirror ascent with only logarithmic dependence on $m$~\citep{2016_Hazan_book}.
However, this static-regret guarantee does not apply directly because $\q_t^*$ varies with $\w_t$, which evolves throughout the optimization process.
A natural alternative is to track the sequence $\{\q_t^*\}_{t=1}^T$ using dynamic-regret techniques~\citep{1998_TBE,2019_TBE_Lu,2021_TBE_Luo}.
Such guarantees, however, depend on the path variation of $\{\q_t^*\}_{t=1}^T$, which can grow linearly with $T$ and render the resulting convergence bound vacuous.
\begin{algorithm}[t]
\caption{\textsc{AFL-BR}}
\label{alg:uafl_br}
\begin{algorithmic}[1]
\STATE Initialize $\w_1\in\R^d$ and $\q_1=\frac{1}{m}\mathbf 1_m$
\FOR{$t=1,2,\ldots,T$}
\STATE \textbf{On each worker $i\in[m]$:}
\STATE \quad Draw $\z_t^{(i)}\sim\P_i$ and compute $\g_{w,t}^{(i)}=\nabla\ell(\w_t;\z_t^{(i)})$ and $\g_{q,t}^{(i)}=\ell(\w_t;\z_t^{(i)})$
\STATE \quad Send $\g_{w,t}^{(i)}$ and $\g_{q,t}^{(i)}$ to the server
\STATE \textbf{On the server:}
\STATE \quad Construct $\g_{q,t}=[\g_{q,t}^{(1)},\ldots,\g_{q,t}^{(m)}]^\top$ and $\g_{w,t}=\sum_{i=1}^m q_{t,i}\g_{w,t}^{(i)}$
\STATE \quad \textbf{if} $t\equiv0\pmod B$ \textbf{then} set $\q_{t+1}=\frac{1}{m}\mathbf 1_m$ \textbf{else} update $\q_{t+1}$ according to~(\ref{eqn:update:q:nonconvex})
\STATE \quad Send $\g_{w,t}$ to all workers
\STATE \textbf{On each worker $i\in[m]$:}
\STATE \quad Receive $\g_{w,t}$ and update $\w_{t+1}=\w_t-\eta_w\g_{w,t}$
\ENDFOR
\STATE \textbf{Output:} $\w_r$, where $r\sim\mathrm{Unif}\{1,\ldots,T\}$
\end{algorithmic}
\end{algorithm}

Following the blockwise analyses~\citep{2020_GDA_minimax,2020_DRFA}, we fix the comparator over short intervals. Specifically, we partition $[T]$ into consecutive blocks $\{\mathcal I_c\}_{c=1}^N$ of length at most $B$, and let $s_c$ denote the first index of $\mathcal I_c$. Then, for every $t\in\mathcal I_c$, we decompose the dual gap as
\begin{equation}\label{eqn:nonconvex_dual_gap_decomp}
\varepsilon_t^q=\underbrace{\phi(\w_t,\q_t^*)-\phi(\w_t,\q_{s_c}^*)}_{A_t}+\underbrace{\phi(\w_t,\q_{s_c}^*)-\phi(\w_t,\q_t)}_{B_t}.
\end{equation}
The first term measures the mismatch caused by fixing the comparator as the model evolves, while the second measures the dual error against this fixed comparator.

By the optimality of $\q_{s_c}^*$ at $\w_{s_c}$ and the $G$-Lipschitzness of $\phi(\cdot,\q)$ under Assumption~\ref{ass:nonconvex_primal_oracle}, we have
\begin{equation*}
\begin{aligned}
A_t&=\phi(\w_t,\q_t^*)-\phi(\w_{s_c},\q_t^*)+\phi(\w_{s_c},\q_t^*)-\phi(\w_t,\q_{s_c}^*)\\
&\leq\phi(\w_t,\q_t^*)-\phi(\w_{s_c},\q_t^*)+\phi(\w_{s_c},\q_{s_c}^*)-\phi(\w_t,\q_{s_c}^*)\leq2G\|\w_t-\w_{s_c}\|_2.
\end{aligned}
\end{equation*}
Hence, the comparator mismatch is controlled by the model movement within the block.
It remains to control $B_t$. Since $\q_{s_c}^*$ is fixed over $\mathcal I_c$, $\sum_{t\in\mathcal I_c}B_t$ has the form of static regret. We therefore update the dual weights using online mirror ascent with KL divergence as
\begin{equation}\label{eqn:update:q:nonconvex}
\q_{t+1}=\argmax_{\q\in\Delta_m}\left\{\eta_q\langle\g_{q,t},\q\rangle-D_{\mathrm{KL}}(\q\|\q_t)\right\},
\end{equation}
where $\g_{q,t}=[g_{q,t}^{(1)},\ldots,g_{q,t}^{(m)}]^\top$ with $g_{q,t}^{(i)}=\ell(\w_t;\z_t^{(i)})$, and $D_{\mathrm{KL}}(\q\|\q_t)=\sum_{i=1}^m q_i\ln(q_i/q_{t,i})$ denotes the KL divergence. The update~(\ref{eqn:update:q:nonconvex}) has the closed form $q_{t+1,i}\propto q_{t,i}\exp(\eta_q g_{q,t}^{(i)}), \forall i\in[m]$.

Moreover, directly running~(\ref{eqn:update:q:nonconvex}) over all $T$ rounds does not provide a uniform blockwise guarantee, since the static-regret bound on each block $\mathcal I_c$ depends on the initial divergence $D_{\mathrm{KL}}(\q_{s_c}^*\|\q_{s_c})$, which is not uniformly bounded. Indeed, $\q_{s_c}$ can assign arbitrarily small mass to the support of the comparator $\q_{s_c}^*$. To resolve this issue, we restart the dual weights as $\q_{s_c}=m^{-1}\mathbf 1_m$ at each block boundary, guaranteeing $D_{\mathrm{KL}}(\q_{s_c}^*\|\q_{s_c})\leq\ln m$ and hence a uniform bound across blocks.

Combining the restarted dual update with stochastic gradient descent on the model yields \textsc{AFL-BR}, summarized in Algorithm~\ref{alg:uafl_br}. At each round, the workers evaluate both stochastic gradients and stochastic losses, while the server aggregates the gradients according to the current weights.

We next establish the convergence guarantee. A further challenge is that standard static-regret bounds depend on the magnitudes of the stochastic losses~\citep{cesa2006prediction}, which are not uniformly bounded here.
To this end, we decompose the stochastic loss into the true loss and noise, and control the noise jointly with the KL stability term through the negative-entropy and log-sum-exp conjugacy.
The sub-Gaussian condition then yields a finite bound without requiring uniform boundedness of either $R_i(\w)$ or $\ell(\w;\z)$, leading to the following guarantee.
\begin{theorem}\label{thm:uafl_br}
Suppose Assumptions~\ref{ass:nonconvex_smooth}--\ref{ass:nonconvex_dual_oracle} hold with $m\geq2$ and $T\geq9$. Let $\Delta_\Phi=\Phi_{1/(2L)}(\w_1)-\Phi_*$. For Algorithm~\ref{alg:uafl_br}, set $B=\lceil\sqrt T\rceil$, $\eta_w=(8LT^{3/4})^{-1}$, and $\eta_q=2\sqrt{\ln m}/(\sigma_qT^{1/4})$. Then
\[
\mathbb E\left[\left\|\nabla\Phi_{1/(2L)}(\w_r)\right\|_2^2\right]
\leq \frac{32L\Delta_\Phi}{T^{1/4}}
+\frac{4G\Gamma_w}{T^{1/4}}
+\frac{16L\sigma_q\sqrt{\ln m}}{T^{1/4}}+\frac{10\Gamma_w^2}{T^{1/2}}
+\frac{\Gamma_w^2}{2T^{3/4}}.
\]
\end{theorem}

\textbf{Remark.}
Theorem~\ref{thm:uafl_br} implies $\E[\|\nabla\Phi_{1/(2L)}(\w_r)\|_2]=O((\log m)^{1/4}T^{-1/8})$ by Jensen's inequality. Hence, obtaining an $\varepsilon$-stationary solution requires $O((\log m)^2\varepsilon^{-8})$ synchronization rounds. 
Compared with the baselines in Table~\ref{tab:AFL-summary}, this reduces the $m$-dependence of the number of synchronization rounds required for convergence from polynomial to logarithmic order.
With $O(dm)$ scalar entries communicated per round, the resulting total communication complexity is $O(dm(\log m)^2\varepsilon^{-8})$.

\textbf{Remark.}
As a byproduct, \textsc{AFL-BR} also yields an $O((\log m)^{1/4}T^{-1/8})$ stationarity guarantee for centralized nonconvex GDRO when all worker-side computations are executed on a single machine.

\section{\textsc{AFL-Com}: AFL with Bidirectional Compression} \label{sec:nonconvex_aflcom_method}
\textsc{AFL-BR} reduces the $m$-dependence of the synchronization-round bound through faster convergence. We further reduce the per-round communication cost via bidirectional compression with $\delta$-approximate compressors, defined below with examples in Appendix~\ref{app:exam_compressors}.
\begin{definition} \label{def:compressor}
	An operator $\C:\R^d\to\R^d$ is a $\delta$-approximate compressor if, for some $\delta\in(0,1]$, $\E[\|\C(\x)-\x\|_2^2]\leq(1-\delta)\|\x\|_2^2$, for all $\x\in\R^d,$ where the expectation is over the randomness of $\C$. 
\end{definition}

\begin{algorithm}[t]
\caption{\textsc{AFL-Com}}
\label{alg:aflcom_br}
\begin{algorithmic}[1]
\STATE Initialize $\w_1\in\R^d$, $\q_1=\frac{1}{m}\mathbf 1_m$, $\widehat\e_1=\mathbf 0_d$, and $\e_1^{(i)}=\mathbf 0_d$ for all $i\in[m]$
\FOR{$t=1,2,\ldots,T$}
\STATE \textbf{On each worker $i\in[m]$:}
\STATE \quad Draw $\z_t^{(i)}\sim\P_i$ and compute $\g_{w,t}^{(i)}=\nabla\ell(\w_t;\z_t^{(i)})$ and $\g_{q,t}^{(i)}=\ell(\w_t;\z_t^{(i)})$
\STATE \quad Compute $\Delta_{w,t}^{(i)}=\mathcal C(\e_t^{(i)}+q_{t,i}\g_{w,t}^{(i)})$ and update $\e_{t+1}^{(i)}=\e_t^{(i)}+q_{t,i}\g_{w,t}^{(i)}-\Delta_{w,t}^{(i)}$
\STATE \quad Send $\Delta_{w,t}^{(i)}$ and $\g_{q,t}^{(i)}$ to the server
\STATE \textbf{On the server:}
\STATE \quad Construct $\g_{q,t}=[\g_{q,t}^{(1)},\ldots,\g_{q,t}^{(m)}]^\top$ and $\hatgb_{w,t}=\sum_{i=1}^m\Delta_{w,t}^{(i)}$
\STATE \quad Compute $\Delta_{w,t}=\mathcal C(\hateb_t+\hatgb_{w,t})$ and update $\widehat\e_{t+1}=\hateb_t+\hatgb_{w,t}-\Delta_{w,t}$
\STATE \quad \textbf{if} $t\equiv0\pmod B$ \textbf{then} set $\q_{t+1}=\frac{1}{m}\mathbf 1_m$ \textbf{else} update $\q_{t+1}$ according to~(\ref{eqn:update:q:nonconvex})
\STATE \quad Broadcast $\Delta_{w,t}$ to all workers and send $q_{t+1,i}$ to worker $i$
\STATE \textbf{On each worker $i\in[m]$:}
\STATE \quad Receive $\Delta_{w,t}$ and $q_{t+1,i}$, and update $\w_{t+1}=\w_t-\eta_w\Delta_{w,t}$
\ENDFOR
\STATE \textbf{Output:} $\w_r$, where $r\sim\mathrm{Unif}\{1,\ldots,T\}$
\end{algorithmic}
\end{algorithm}

\subsection{Bidirectional Compression and Guarantees under General Compressors}\label{sec:aflcom_general}
In each round of \textsc{AFL-BR}, workers upload high-dimensional local gradients and scalar losses, while the server broadcasts the aggregated gradient. 
Since gradient transmission dominates the communication cost, we compress these model-update messages bidirectionally and use EF~\citep{EF_ICML19} to control the compression errors.
A natural approach is to apply EF independently to each worker's raw local gradient, i.e., worker $i$ transmits $\Delta_{w,t}^{(i)}=\C(\e_t^{(i)}+\g_{w,t}^{(i)})$ and accumulates the residual $\e_{t+1}^{(i)}=\e_t^{(i)}+\g_{w,t}^{(i)}-\Delta_{w,t}^{(i)}$.
The server then aggregates the compressed messages using the current dual weights.
However, since the weights are applied \emph{outside} the EF recursion, the accumulated uplink error $\sum_{s=1}^t\sum_{i=1}^m q_{s,i}(\g_{w,s}^{(i)}-\Delta_{w,s}^{(i)})$ is not directly controlled by the worker-side EF guarantee, which only bounds the unweighted residual $\sum_{s=1}^t(\g_{w,s}^{(i)}-\Delta_{w,s}^{(i)})$. This mismatch leads to a loose bound on the weighted global error.

We address the mismatch by applying EF to the dual-weighted gradients $q_{t,i}\g_{w,t}^{(i)}$. To this end, the server sends $q_{t,i}$ to worker $i$ at the end of round $t-1$. 
This incurs a communication cost of $m$ floats per round, which is negligible compared with communicating high-dimensional gradients.
Specifically, worker $i$ sends $\Delta_{w,t}^{(i)}=\mathcal C(\e_t^{(i)}+q_{t,i}\g_{w,t}^{(i)})$ and updates $\e_{t+1}^{(i)}=\e_t^{(i)}+q_{t,i}\g_{w,t}^{(i)}-\Delta_{w,t}^{(i)}$. 
The server aggregates $\hatgb_{w,t}=\sum_{i=1}^m\Delta_{w,t}^{(i)}$ and applies EF again on the downlink by broadcasting $\Delta_{w,t}=\mathcal C(\hateb_t+\hatgb_{w,t})$ and updating $\widehat\e_{t+1}=\hateb_t+\hatgb_{w,t}-\Delta_{w,t}$. 
Each worker then updates $\w_{t+1}=\w_t-\eta_w\Delta_{w,t}$, while the dual weights follow the same blockwise-restarted mirror-ascent update as in \textsc{AFL-BR}.
We present the resulting procedure in Algorithm~\ref{alg:aflcom_br}, termed \textsc{AFL-Com}, and establish its convergence under general $\delta$-approximate compressors as follows.
\begin{theorem}\label{thm:aflcom_br_general}
Suppose Assumptions~\ref{ass:nonconvex_smooth}--\ref{ass:nonconvex_dual_oracle} hold with $m\geq2$ and $T\geq9$. Let $\Delta_\Phi=\Phi_{1/(2L)}(\w_1)-\Phi_*$. For Algorithm~\ref{alg:aflcom_br}, set $B=\lceil\sqrt T\rceil$, $\eta_w=(8LT^{3/4})^{-1}$, and $\eta_q=2\sqrt{\ln m}/(\sigma_qT^{1/4})$. If $\mathcal C$ is a general $\delta$-approximate compressor, then
\begin{equation}\label{eqn:aflcom_br_general_rate_new}
\begin{aligned}
\mathbb E\left[\left\|\nabla\Phi_{1/(2L)}(\w_r)\right\|_2^2\right]
\leq{}&\frac{32L\Delta_\Phi}{T^{1/4}}+\frac{4G\Gamma_w(16+4\delta-\delta^2)}{\delta^2T^{1/4}}+\frac{16L\sigma_q\sqrt{\ln m}}{T^{1/4}}+\frac{10\Gamma_w^2}{T^{1/2}}\\
&+\frac{3G\Gamma_w(8+2\delta-\delta^2)}{\delta^2T^{3/4}}+\frac{\Gamma_w^2}{2T^{3/4}}.
\end{aligned}
\end{equation}
\end{theorem}
\textbf{Remark.}
Theorem~\ref{thm:aflcom_br_general} establishes a stationarity bound of $O((\delta^{-1}+(\log m)^{1/4})T^{-1/8})$. Hence, choosing $\delta=\Omega((\log m)^{-1/4})$ preserves the $O((\log m)^{1/4}T^{-1/8})$ rate of \textsc{AFL-BR} while reducing the per-round communication cost. 
For Top-$k$ with $\delta=k/d$, taking $k=\Theta(d(\log m)^{-1/4})$ reduces the per-round communication cost to $O(dm(\log m)^{-1/4})$, yielding a total communication complexity of $O(dm(\log m)^{7/4}\varepsilon^{-8})$, a factor of $(\log m)^{1/4}$ lower than that of \textsc{AFL-BR}.

\subsection{Sharper Guarantee for Additive-and-Idempotent Compressors}\label{sec:aflcom_ai}
For fixed-weight distributed nonconvex optimization, i.e., $\q=m^{-1}\mathbf{1}_m$ in~(\ref{eq:gdro}), the compression-dependent lower bound on expected squared gradient norm scales as $\Omega(\delta^{-1})$~\citep{FCC-NIPS22}. 
This suggests that the $O(\delta^{-2})$ dependence in Theorem~\ref{thm:aflcom_br_general} may be improvable and raises the question of whether sharper control of compression errors can improve the $\delta$-dependence. We answer this question affirmatively for additive-and-idempotent compressors with shared randomness (SR).
\begin{definition} \label{def:compressor:add-idem}
A randomized compressor satisfying Definition~\ref{def:compressor} is written as $\C^{(\xi)}(\cdot)$, where $\xi$ denotes its internal randomness.
For every realization $\xi$, we define the following two properties:
\begin{compactenum}
    \item $\C$ is \emph{additive} if $\C^{(\xi)}\left(\sum_{i=1}^{n}\u_i\right)=\sum_{i=1}^{n}\C^{(\xi)}(\u_i)$ for any $n\in\N$ and $\u_1,\dots,\u_n\in\R^d$.
    \item $\C$ is \emph{idempotent} if $\C^{(\xi)}\bigl(\C^{(\xi)}(\x)\bigr)=\C^{(\xi)}(\x)$ for any $\x\in\R^d$.
\end{compactenum}
We denote the class of compressors satisfying both properties by $\C_{\mathrm{AI}}$.
\end{definition}
Examples of compressors in $\C_{\mathrm{AI}}$ are detailed in Appendix~\ref{app:exam_compressors}. In particular, unscaled Rand-$k$ is in $\C_{\mathrm{AI}}$ with $\delta=k/d$.
For randomized compressors, SR means that the server and all workers use the same compressor realization in each round. 
We note that SR can be implemented without per-round coordination or additional per-round communication, with details provided in Appendix~\ref{appendix:SR}.

\noindent\textbf{Key observation.}
When \textsc{AFL-Com} uses a compressor $\C\in\C_{\mathrm{AI}}$ with SR, the \emph{downlink compression introduces no error}. Indeed, for all $t\in[T]$, the uplink aggregate $\hatgb_{w,t}$ satisfies
\begin{align*}
\C^{(\xi_t)}(\hatgb_{w,t})
=& \C^{(\xi_t)}\left(\sum_{i=1}^m\C^{(\xi_t)}\left(\e_t^{(i)}+q_{t,i}\g_{w,t}^{(i)}\right)\right) 
\overset{\text{(additivity)}}{=} \C^{(\xi_t)}\left(\C^{(\xi_t)}\left(\sum_{i=1}^m \e_t^{(i)}+\sum_{i=1}^m q_{t,i}\g_{w,t}^{(i)}\right)\right) \\
\overset{\text{(idempotence)}}{=}&\C^{(\xi_t)}\left(\sum_{i=1}^m \e_t^{(i)}+\sum_{i=1}^m q_{t,i}\g_{w,t}^{(i)}\right)
\overset{\text{(additivity)}}{=} \sum_{i=1}^m\C^{(\xi_t)}\left(\e_t^{(i)}+q_{t,i}\g_{w,t}^{(i)}\right) = \hatgb_{w,t}.
\end{align*}
Since $\hateb_1=\mathbf0_d$, the server-side EF recursion yields $\hateb_t=\mathbf0_d$ for all $t\in[T]$. Hence, the downlink introduces no additional compression residual, as formalized in Lemma~\ref{lem:compression_residuals}. This sharper control improves the dependence on $\delta$, leading to the following convergence guarantee.
\begin{theorem}\label{thm:aflcom_br_ai}
Suppose Assumptions~\ref{ass:nonconvex_smooth}--\ref{ass:nonconvex_dual_oracle} hold with $m\geq2$ and $T\geq9$. Let $\Delta_\Phi=\Phi_{1/(2L)}(\w_1)-\Phi_*$. For Algorithm~\ref{alg:aflcom_br}, set $B=\lceil\sqrt T\rceil$, $\eta_w=(8LT^{3/4})^{-1}$, and $\eta_q=2\sqrt{\ln m}/(\sigma_qT^{1/4})$. If $\mathcal C\in\mathcal C_{\mathrm{AI}}$ and SR is used, then
\begin{equation*}
\mathbb E\left[\left\|\nabla\Phi_{1/(2L)}(\w_r)\right\|_2^2\right]
\leq
\frac{32L\Delta_\Phi}{T^{1/4}}
+\frac{16G\Gamma_w}{\delta T^{1/4}}
+\frac{16L\sigma_q\sqrt{\ln m}}{T^{1/4}}
+\frac{10\Gamma_w^2}{T^{1/2}}+\frac{6G\Gamma_w}{\delta T^{3/4}}
+\frac{\Gamma_w^2}{2T^{3/4}}.
\end{equation*}
\end{theorem}
\textbf{Remark.}
Theorem~\ref{thm:aflcom_br_ai} implies a stationarity bound of $O((\delta^{-1/2}+(\log m)^{1/4})T^{-1/8})$, improving the dependence on $\delta$ from $\delta^{-1}$ in Theorem~\ref{thm:aflcom_br_general} to $\delta^{-1/2}$. Hence, choosing $\delta=\Omega((\log m)^{-1/2})$ preserves the $O((\log m)^{1/4}T^{-1/8})$ rate of \textsc{AFL-BR}. 
For Rand-$k$ with SR, taking $k=\Theta(d(\log m)^{-1/2})$ reduces the communication cost per round to $O(dm(\log m)^{-1/2})$ and yields a communication complexity of $O(dm(\log m)^{3/2}\varepsilon^{-8})$, a factor of $(\log m)^{1/2}$ lower than that of \textsc{AFL-BR}.

\section{Experiments}\label{sec:experiments}
We evaluate the two components of our communication-efficiency gains, namely, reducing the number of synchronization rounds and reducing the per-round communication cost through compression.

\subsection{Scaling with the Number of Workers} \label{sec:exp_scaling}
We first examine how the number of workers affects optimization efficiency. We use FEMNIST~\citep{2018_LEAF}, which provides naturally heterogeneous writer-level partitions with $62$ output classes, and consider $m\in\{20,50,100,500,1000\}$. 
For each $m$, we rank writers by their number of local samples and select the top-$m$ writers, with each writer's data randomly split 80/20 into training and test sets.

\begin{wrapfigure}{r}{0.43\columnwidth}
\centering
\includegraphics[width=\linewidth]{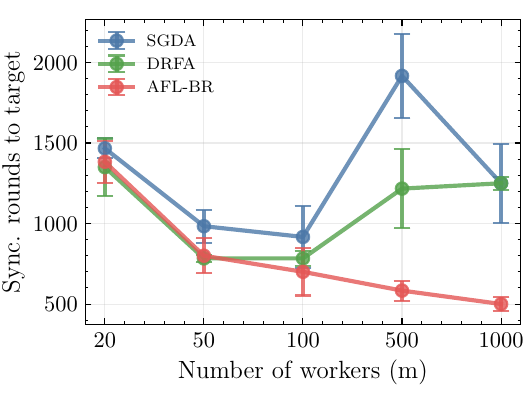}
\caption{Synchronization rounds versus $m$ on FEMNIST.}
\label{fig:rounds_vs_m}
\vspace{-1em}
\end{wrapfigure}
We compare \textsc{AFL-BR} with \textsc{SGDA} and \textsc{DRFA} under full participation, with \textsc{DRFA} performing three local updates per synchronization. All methods train a two-layer CNN from scratch, with more details provided in Appendix~\ref{app:exp_scaling}.
For each $m$, we choose a shared target worst-worker test accuracy for all methods and record the first checkpoint reaching that target. We report the mean over five runs, with error bars showing one standard deviation.
As shown in Figure~\ref{fig:rounds_vs_m}, \textsc{AFL-BR} and \textsc{DRFA} are comparable for $m=20$ and $m=50$, while \textsc{AFL-BR} reaches the target in fewer synchronization rounds than both baselines for $m=100$, $500$, and $1000$.
These large-$m$ results are consistent with the improved $m$-dependence in the number of synchronization rounds achieved by \textsc{AFL-BR}, demonstrating its efficiency as the federation scales.

\subsection{Communication Efficiency with Compression}\label{sec:exp:compression}
We evaluate \textsc{AFL-BR} and \textsc{AFL-Com} on Fashion-MNIST~\citep{xiao2017fashionmnist} and CIFAR-10~\citep{CIFAR-10} with $m=10$ workers and label-skewed data. We instantiate Algorithm~\ref{alg:aflcom_br} with Top-$k$ and Rand-$k$, denoted by \textsc{AFL-Com}(T) and \textsc{AFL-Com}(R), retaining $30\%$ and $10\%$ of the coordinates, respectively. The Rand-$k$ variant uses shared randomness across all workers and the server.
The baselines include \textsc{AFL}~\citep{2019AFL}, equivalent here to applying \textsc{SGDA} to problem~(\ref{eq:gdro}), and \textsc{DRFA}~\citep{2020_DRFA}.
We also include \textsc{FedAvg}~\citep{2017_FL_McMahan} as a standard FL baseline that minimizes the average local loss. \textsc{FedAvg} and \textsc{DRFA} perform three local updates per synchronization, whereas \textsc{AFL}, \textsc{AFL-BR}, and \textsc{AFL-Com} perform one model update. Further details and results are provided in Appendix~\ref{app:exp_compression}.
Figure~\ref{fig:nonconvex_acc} reports worst-worker test accuracy against update rounds, synchronization rounds, and cumulative communication cost. Results are averaged over five runs, with shaded regions indicating one standard deviation.
\begin{figure*}[t]
\centering
\includegraphics[width=\textwidth]{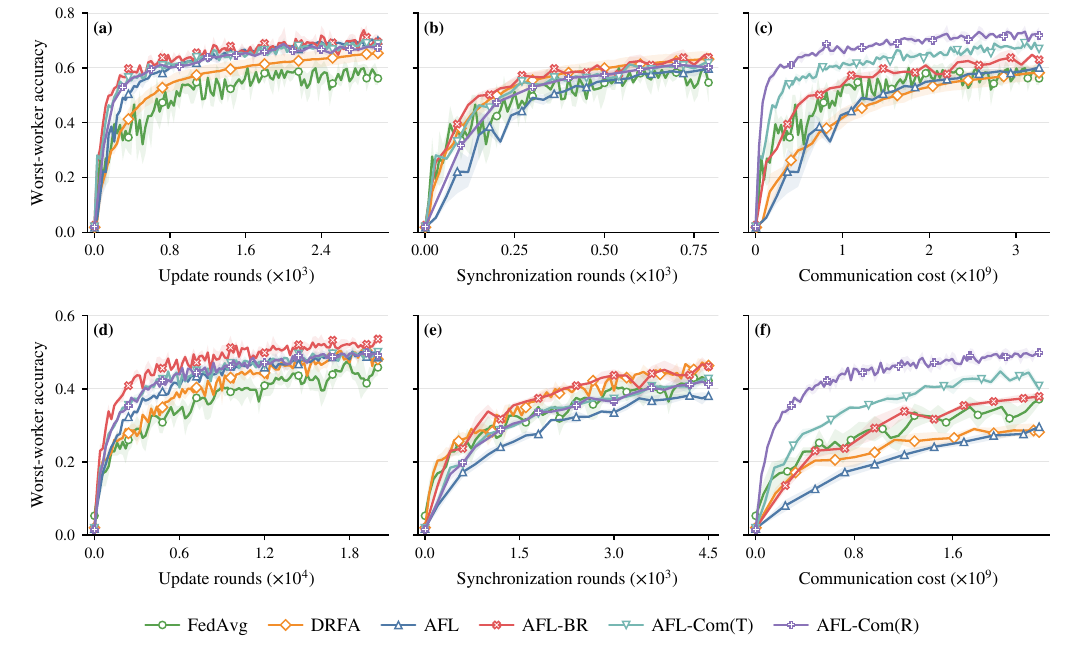}
\caption{\textbf{Worst-worker test accuracy.} The top and bottom rows correspond to Fashion-MNIST and CIFAR-10, respectively. From left to right, the columns report performance against update rounds, synchronization rounds, and cumulative communication cost.}
\label{fig:nonconvex_acc}
\end{figure*}

As shown in Figure~\ref{fig:nonconvex_acc}(a) and (d), the AFL methods generally achieve higher worst-worker test accuracy than \textsc{FedAvg}, consistent with their objective of minimizing the worst-worker risk. Among the AFL methods, \textsc{AFL-BR} reaches a given worst-worker accuracy in fewer update rounds than \textsc{AFL} and \textsc{DRFA}, indicating faster convergence, while \textsc{AFL-Com}(T) and \textsc{AFL-Com}(R) closely track \textsc{AFL-BR} despite compression.
Panels (b) and (e) of Figure~\ref{fig:nonconvex_acc} show that \textsc{AFL-BR} reaches a given accuracy in fewer synchronization rounds than \textsc{AFL} and remains competitive with or better than \textsc{DRFA}. 
Despite a less favorable synchronization-round bound than its fully synchronized specialization, \textsc{DRFA} improves over \textsc{AFL} in practice through multiple local updates. Moreover, \textsc{AFL-BR} achieves comparable or better efficiency through faster convergence with only one update per synchronization.

We finally compare the performance against cumulative communication cost. Each Top-$k$ message costs $k+k\lceil\log_2 d\rceil/32$ float-equivalents, accounting for values and indices, while Rand-$k$ with SR costs only $k$, as the coordinate indices are reconstructed locally using SR.
Panels (c) and (f) of Figure~\ref{fig:nonconvex_acc} show that both \textsc{AFL-Com} variants attain comparable worst-worker accuracy at substantially lower communication cost than the uncompressed AFL baselines. 
\textsc{AFL-Com}(R) is the most communication-efficient on both datasets, consistent with our sharper analysis for additive-and-idempotent compressors with SR, where the downlink compression residual vanishes.

Overall, panels (a), (b), (d), and (e) of Figure~\ref{fig:nonconvex_acc} illustrate the benefit of faster convergence, with \textsc{AFL-BR} requiring fewer update and synchronization rounds to attain a given accuracy, while panels (c) and (f) show that \textsc{AFL-Com} further reduces the communication cost through compression.

\section{Conclusion}
We improve the communication efficiency of AFL by reducing both the number of synchronization rounds required for convergence and the communication cost per round. 
We first propose \textsc{AFL-BR}, which reduces the worker dependence of existing guarantees from polynomial to logarithmic order. 
Building on it, we develop \textsc{AFL-Com} by introducing bidirectional compression with EF for dual-weighted local gradients. 
We establish its convergence under general $\delta$-approximate compressors and further obtain a sharper dependence on $\delta$ for additive-and-idempotent compressors with SR. 
Experiments suggest substantial communication savings with competitive worst-worker performance.

\bibliographystyle{iclr2027_conference}
\bibliography{ref}

\appendix
\section{Additional Related Work} \label{app:related_work}
For convex AFL, \citet{2019AFL} achieve an $O(T^{-1/2})$ optimization-error rate. \textsc{DRFA}~\citep{2020_DRFA} uses local updates to reduce the number of synchronization rounds to $O(T^{3/4})$, but slows the optimization-error rate to $O(T^{-3/8})$. Consequently, both methods require $O(\varepsilon^{-2})$ synchronization rounds to reach an $\varepsilon$-accurate solution. Other convex variants incorporate worker-drift correction~\citep{2025_GDRO_dis_DRDM}, use local updates to improve empirical communication efficiency without convergence guarantees~\citep{2021CEAFA_noconvergence}, or adopt alternative communication models based on pretrained predictor ensembles~\citep{2020_FedBoost_Ensemble}. 
A complementary line of work studies constrained, regularized, or decentralized variants of AFL and federated GDRO~\citep{2022_GDRO_dec,2023_arxiv_Scaff_PD,2024_NIPS_FGDRO_Yang,2023_GDRO_dec_cp}.

\section{Additional Discussions}  \label{appendix:discussion}
\subsection{Examples of Compressors}\label{app:exam_compressors}
We provide several representative compressors considered in this paper. 
More extensive collections of biased and unbiased compressors, together with their contraction or variance parameters, can be found in~\citet{2022_Com_example,2023_JMLR_com_example} and the references therein.

We first give standard examples satisfying Definition~\ref{def:compressor}.
\begin{compactitem}
\item \emph{Top-$k$ sparsification}~\citep{2018_SparSGD_toprandk} is defined by $\C_{\mathrm{top}\text{-}k}(\x)=\sum_{j\in S_k(\x)}x_j\e_j$, where $S_k(\x)$ indexes the $k$ coordinates of $\x$ with the largest absolute values. Since the retained coordinates contain at least a $k/d$ fraction of the squared Euclidean norm, $\|\C_{\mathrm{top}\text{-}k}(\x)-\x\|_2^2\leq (1-k/d)\|\x\|_2^2.$
Hence, Definition~\ref{def:compressor} holds with $\delta=k/d$. The support $S_k(\x)$ depends on the input, so Top-$k$ is nonlinear and is not additive in general.

\item \emph{Scaled-sign compression}~\citep{EF_ICML19} is defined by $\C_{\mathrm{sign}}(\x)=\frac{\|\x\|_1}{d}\operatorname{sign}(\x),$ with $\operatorname{sign}(0)=0$. A direct calculation gives $\|\C_{\mathrm{sign}}(\x)-\x\|_2^2\leq\|\x\|_2^2-\|\x\|_1^2/d\leq\left(1-1/d\right)\|\x\|_2^2,$ where the second inequality follows from $\|\x\|_1\geq\|\x\|_2$. Thus, it is a $\delta$-approximate compressor with $\delta=1/d$.
\end{compactitem}

The class $\C_{\mathrm{AI}}$ additionally requires additivity and idempotence for every fixed realization of the compressor randomness. Two useful examples are randomized projection operators.
\begin{compactitem}
\item \emph{Unscaled Rand-$k$ sparsification}~\citep{2018_SparSGD_toprandk}.
Let $S\subseteq[d]$ be sampled uniformly from all subsets of cardinality $k$, and define $\C_{\mathrm{rand}\text{-}k}^{(S)}(\x)=P_S\x=\sum_{j\in S}x_j\e_j.$
For every fixed $S$, $P_S$ is linear and satisfies $P_S^2=P_S$. Hence, the compressor is additive and idempotent pathwise. Moreover, $\E_S[\|P_S\x-\x\|_2^2]=(1-k/d)\|\x\|_2^2,$ so Definition~\ref{def:compressor} holds with $\delta=k/d$.

We emphasize that this is the \emph{unscaled} contractive version of Rand-$k$. The commonly used unbiased scaling $(d/k)P_S\x$ is not idempotent unless $k=d$. Under SR, all workers and the server use the same subset $S$ within a communication round, so the coordinate indices need not be transmitted.

\item \emph{Random rank-$r$ orthogonal projection}~\citep{vershynin2018high}.
Let $U\in\R^{d\times r}$ have orthonormal columns spanning a uniformly random $r$-dimensional subspace, and define $\C_{\mathrm{proj}}^{(U)}(\x)=UU^\top\x.$
For every fixed $U$, the matrix $P_U=UU^\top$ is linear and satisfies $P_U^2=P_U$. Therefore, the compressor is additive and idempotent. By rotational symmetry, $\E_U[P_U]=\frac{r}{d}I_d,$
and consequently $\E_U[\|UU^\top\x-\x\|_2^2]=(1-r/d)\|\x\|_2^2.$ Thus, Definition~\ref{def:compressor} holds with $\delta=r/d$.
With SR, the subspace $U$ can be reconstructed locally from the common random seed, so only the $r$ coefficients $U^\top\x$ need to be transmitted. The receiver can then reconstruct $UU^\top\x$ locally.
\end{compactitem}

\subsection{Implementation of Shared Randomness} \label{appendix:SR}
Shared randomness (SR) requires the server and all workers to use the same compressor realization in each communication round. 
In practice, SR can be implemented without per-round coordination.
Before training, the server and all workers synchronize a single random seed.
Then, in each round, they locally generate a common realization using a fixed deterministic rule~\citep{2020_TIT_SR_app_1} or synchronized pseudorandom number generators~\citep{2022_ICML_SR_app}. 
Similar SR mechanisms have been used in communication-efficient distributed learning, for example, by~\citet{2026_TSP_SR_app}. 
Therefore, SR introduces no per-round coordination or additional per-round communication.
The only overhead is the initial transmission of a scalar seed, which is negligible compared with transmitting high-dimensional model updates.

\section{Details for Table~\ref{tab:AFL-summary}} \label{app:table_details}
This section explains two technical details used in Table~\ref{tab:AFL-summary}. 

\subsection{Specialization of SGDA to AFL}\label{app:sgda_hidden_m}
The SGDA guarantee in~\cite[Theorem~4.9]{2020_GDA_minimax} applies to general nonconvex--concave minimax problems. We specialize this result to the centralized GDRO objective $\phi(\w,\q)=\sum_{i=1}^m q_iR_i(\w)$ with $\q\in\Delta_m$ and make explicit the dependence hidden in its Euclidean joint smoothness and stochastic-oracle variance parameters. Unlike the DRFA analysis discussed in Appendix~\ref{app:drfa_refined_gamma}, this specialization does not require a uniform bound on $\|\mathbf R(\w)\|_2$. Its dependence on $m$ instead arises from the Euclidean geometry of the $m$-dimensional dual weights.

\paragraph{Primal Lipschitzness and joint smoothness.} Let $G_\phi$ denote the uniform Lipschitz constant of $\phi(\cdot,\q)$ and let $L_{\mathrm{joint}}$ denote the Euclidean joint smoothness constant of $\phi$, corresponding to the parameter denoted by $\ell$ in~\cite[Assumption~4.6]{2020_GDA_minimax}. By Assumption~\ref{ass:nonconvex_primal_oracle}, for every $\q\in\Delta_m$,
\begin{equation*}\label{eqn:sgda_primal_lipschitz}
\|\nabla_{\w}\phi(\w,\q)\|_2=\left\|\sum_{i=1}^m q_i\nabla R_i(\w)\right\|_2\leq\sum_{i=1}^m q_i\|\nabla R_i(\w)\|_2\leq G.
\end{equation*}
Thus, $\phi(\cdot,\q)$ is $G$-Lipschitz uniformly over $\q\in\Delta_m$, and one may take $G_\phi=G$.

Let $\mathbf R(\w)=(R_1(\w),\ldots,R_m(\w))^\top$. The joint gradient is
\begin{equation}\label{eqn:sgda_joint_gradient}
\nabla\phi(\w,\q)=
\begin{bmatrix}
\sum_{i=1}^m q_i\nabla R_i(\w)\\
\mathbf R(\w)
\end{bmatrix}.
\end{equation}
Since $\|\nabla R_i(\w)\|_2\leq G$ under Assumption~\ref{ass:nonconvex_primal_oracle}, each $R_i$ is $G$-Lipschitz. Therefore,
\begin{equation*}
\|\mathbf R(\w)-\mathbf R(\w')\|_2=\left(\sum_{i=1}^m|R_i(\w)-R_i(\w')|^2\right)^{1/2}\leq G\sqrt m\,\|\w-\w'\|_2.
\end{equation*}
For the primal component of~(\ref{eqn:sgda_joint_gradient}), Assumptions~\ref{ass:nonconvex_smooth} and~\ref{ass:nonconvex_primal_oracle} imply
\begin{equation*}
\begin{aligned}
&\left\|\sum_{i=1}^m q_i\nabla R_i(\w)-\sum_{i=1}^m q_i'\nabla R_i(\w')\right\|_2\\
&\leq\left\|\sum_{i=1}^m q_i\bigl(\nabla R_i(\w)-\nabla R_i(\w')\bigr)\right\|_2+\left\|\sum_{i=1}^m(q_i-q_i')\nabla R_i(\w')\right\|_2\\
&\leq L\|\w-\w'\|_2+G\|\q-\q'\|_1\\
&\leq L\|\w-\w'\|_2+G\sqrt m\,\|\q-\q'\|_2.
\end{aligned}
\end{equation*}
Combining the two inequalities gives
\begin{equation*}\label{eqn:sgda_joint_smoothness}
\|\nabla\phi(\w,\q)-\nabla\phi(\w',\q')\|_2^2\leq2\bigl(L^2+mG^2\bigr)\left(\|\w-\w'\|_2^2+\|\q-\q'\|_2^2\right).
\end{equation*}
Hence, one may take $L_{\mathrm{joint}}=\sqrt{2(L^2+mG^2)}=O(\sqrt m)$.

We next show that the $\sqrt m$ dependence of $L_{\mathrm{joint}}$ cannot generally be removed under the Euclidean geometry used by SGDA. Since $\nabla_{\q}\phi(\w,\q)=\mathbf R(\w)$, joint $L_{\mathrm{joint}}$-smoothness implies
\begin{align*}
    &\|\mathbf R(\w)-\mathbf R(\w')\|_2=\|\nabla_{\q}\phi(\w,\q)-\nabla_{\q}\phi(\w',\q)\|_2\\
    \leq&\|\nabla\phi(\w,\q)-\nabla\phi(\w',\q)\|_2\leq L_{\mathrm{joint}}\|\w-\w'\|_2,
\end{align*}
and therefore
\begin{equation*}\label{eqn:sgda_joint_smoothness_lower_general}
L_{\mathrm{joint}}\geq\sup_{\w\neq\w'}\frac{\|\mathbf R(\w)-\mathbf R(\w')\|_2}{\|\w-\w'\|_2}.
\end{equation*}
To see that the right-hand side can be of order $\sqrt m$, let $\mathbf a\in\mathbb R^d$ satisfy $\|\mathbf a\|_2=1$ and define $R_i(\w)=\frac{1}{2}+\frac{1}{2}\tanh\left(2G\langle\mathbf a,\w\rangle\right)$ for $i\in[m]$.
Then $R_i(\w)\in(0,1), \ \nabla R_i(\w)=G\operatorname{sech}^2\left(2G\langle\mathbf a,\w\rangle\right)\mathbf a,$ and therefore $\|\nabla R_i(\w)\|_2\leq G$. Moreover, $\|\nabla^2R_i(\w)\|_2\leq4G^2$, so the smoothness constant of each local objective is independent of $m$. Taking $\w'=\mathbf0_d$ and $\w=h\mathbf a$ gives
\begin{equation*}\label{eqn:sgda_joint_smoothness_lower_example}
\frac{\|\mathbf R(h\mathbf a)-\mathbf R(\mathbf0_d)\|_2}{\|h\mathbf a\|_2}=\sqrt m\,\frac{|\tanh(2Gh)|}{2|h|}\longrightarrow G\sqrt m\qquad\text{as }h\to0.
\end{equation*}
Thus, $L_{\mathrm{joint}}=\Omega(\sqrt m)$ in the worst case. Together with the upper bound $L_{\mathrm{joint}}=O(\sqrt m)$, this shows that $L_{\mathrm{joint}}=\Theta(\sqrt m)$ even when all local smoothness and gradient bounds are independent of $m$.

\paragraph{Variance of the joint stochastic oracle.} At a centralized SGDA iteration, consider the unbiased joint stochastic oracle
\begin{equation*}\label{eqn:sgda_joint_oracle}
\widehat{\nabla\phi}(\w,\q;\boldsymbol{\z})=
\begin{bmatrix}
\sum_{i=1}^m q_i\nabla\ell(\w;\z^{(i)})\\
\widehat{\mathbf R}(\w;\boldsymbol{\z})
\end{bmatrix},
\end{equation*}
where $\widehat{\mathbf R}(\w;\boldsymbol{\z})=[\ell(\w;\z^{(1)}),\ldots,\ell(\w;\z^{(m)})]^\top$. By convexity of the squared norm and Assumption~\ref{ass:nonconvex_primal_oracle},
\begin{equation*}
\mathbb E_{\boldsymbol{\z}}\left[\left\|\sum_{i=1}^m q_i\bigl(\nabla\ell(\w;\z^{(i)})-\nabla R_i(\w)\bigr)\right\|_2^2\right]\leq\sum_{i=1}^m q_i\mathbb E_{\z^{(i)}}\left[\|\nabla\ell(\w;\z^{(i)})-\nabla R_i(\w)\|_2^2\right]\leq\sigma_w^2.
\end{equation*}
No independence across workers is required for this inequality. Moreover, Assumption~\ref{ass:nonconvex_dual_oracle} implies $\mathbb E_{\z^{(i)}\sim\P_i}[(\ell(\w;\z^{(i)})-R_i(\w))^2]\leq\sigma_q^2$ for every $i\in[m]$. Hence,
\begin{equation*}
\mathbb E_{\boldsymbol{\z}}\left[\left\|\widehat{\mathbf R}(\w;\boldsymbol{\z})-\mathbf R(\w)\right\|_2^2\right]=\sum_{i=1}^m\mathbb E_{\z^{(i)}\sim\P_i}\left[(\ell(\w;\z^{(i)})-R_i(\w))^2\right]\leq m\sigma_q^2.
\end{equation*}
Consequently, the variance parameter $\sigma_{\mathrm{joint}}^2$ of SGDA can be chosen as $\sigma_{\mathrm{joint}}^2\leq\sigma_w^2+m\sigma_q^2.$

\paragraph{Centralized iteration complexity.} Using the notation above, the stochastic joint-oracle complexity in~\cite[Theorem~4.9]{2020_GDA_minimax} can be written as
\begin{equation}\label{eqn:sgda_complexity_general}
N_{\mathrm{SGDA}}(\varepsilon)=O\left(\left(\frac{L_{\mathrm{joint}}^3(G_\phi^2+\sigma_{\mathrm{joint}}^2)D_\Delta^2\widehat\Delta_\Phi}{\varepsilon^6}+\frac{L_{\mathrm{joint}}^3D_\Delta^2\widehat\Delta_0}{\varepsilon^4}\right)\max\left\{1,\frac{\sigma_{\mathrm{joint}}^2}{\varepsilon^2}\right\}\right),
\end{equation}
where $D_\Delta$ is the Euclidean diameter of $\Delta_m$, $\widehat\Delta_\Phi=\Phi_{1/(2L_{\mathrm{joint}})}(\w_0)-\inf_{\w}\Phi_{1/(2L_{\mathrm{joint}})}(\w),$ and $\widehat\Delta_0=\Phi(\w_0)-\phi(\w_0,\q_0)$ is the initial dual gap. Since $D_\Delta^2\leq2$, $L_{\mathrm{joint}}^3=O(m^{3/2})$ and $\sigma_{\mathrm{joint}}^2=O(m)$,~(\ref{eqn:sgda_complexity_general}) therefore gives $N_{\mathrm{SGDA}}(\varepsilon)=O\left(m^{7/2}\varepsilon^{-8}\right).$ 

The above stationarity criterion is expressed using $\nabla\Phi_{1/(2L_{\mathrm{joint}})}$, whereas the nonconvex results in this paper use $\nabla\Phi_{1/(2L)}$. 
The following comparison shows that the SGDA stationarity guarantee implies our stationarity criterion up to an absolute constant.
\begin{lemma}\label{lem:moreau_parameter_comparison}
Let $f:\mathbb R^d\to\mathbb R$ be $\rho$-weakly convex. For any $0<\mu\leq\lambda<1/\rho$, the gradients of its Moreau envelopes satisfy $\|\nabla f_\lambda(\x)\|_2\leq\frac{1-\rho\mu}{1-\rho\lambda}\|\nabla f_\mu(\x)\|_2$ for all $\x\in\mathbb R^d$.
\end{lemma}

By Lemma~\ref{lem:Phi_weakly_convex}, $\Phi$ is $L$-weakly convex. Since $L_{\mathrm{joint}}=\sqrt{2(L^2+mG^2)}\geq L$, we may apply Lemma~\ref{lem:moreau_parameter_comparison} with $\rho=L$, $\mu=1/(2L_{\mathrm{joint}})$, and $\lambda=1/(2L)$. This gives
\begin{equation*}\label{eqn:sgda_moreau_parameter_transfer}
\begin{aligned}
\left\|\nabla\Phi_{1/(2L)}(\w)\right\|_2
&\leq\frac{1-L/(2L_{\mathrm{joint}})}{1-L/(2L)}\left\|\nabla\Phi_{1/(2L_{\mathrm{joint}})}(\w)\right\|_2\\
&=\left(2-\frac{L}{L_{\mathrm{joint}}}\right)\left\|\nabla\Phi_{1/(2L_{\mathrm{joint}})}(\w)\right\|_2\\
&\leq2\left\|\nabla\Phi_{1/(2L_{\mathrm{joint}})}(\w)\right\|_2.
\end{aligned}
\end{equation*}
Therefore, running SGDA until $\mathbb E[\|\nabla\Phi_{1/(2L_{\mathrm{joint}})}(\w_r)\|_2]\leq\varepsilon/2$ is sufficient to guarantee $\mathbb E[\|\nabla\Phi_{1/(2L)}(\w_r)\|_2]\leq\varepsilon$. Replacing $\varepsilon$ by $\varepsilon/2$ changes only an absolute constant.

Equivalently, after $T$ centralized SGDA iterations, $\mathbb E\left[\left\|\nabla\Phi_{1/(2L)}(\w_r)\right\|_2\right]=O\left(m^{7/16}T^{-1/8}\right).$
When SGDA is implemented in the parameter-server setting, each iteration requires one communication round to aggregate the $m$ local primal gradients and construct the $m$-dimensional stochastic loss vector. Since each round communicates $O(dm)$ entries, the resulting communication complexity is $O\left(dm^{9/2}\varepsilon^{-8}\right).$ 
The above rate and the communication complexity are the SGDA entries reported in Table~\ref{tab:AFL-summary}.

\subsection{Refined DRFA Bounds with Partial Participation}\label{app:drfa_refined_gamma}

We make explicit the worker and participation dependence of the nonconvex DRFA analysis in~\citet{2020_DRFA}. Their notation uses $N$ for the total number of workers and $m$ for the sampling size. Here, $m$ denotes the total number of workers, $b$ the sampling size, and $p=b/m$. We additionally assume $|R_i(\w)|\leq G_R$, with $G_R,\sigma_q,L,G,$ and $\sigma_w$ independent of $m$ and $p$.

\paragraph{Dual oracle under partial participation.} Let $\mathbf R(\w)=(R_1(\w),\ldots,R_m(\w))^\top=\nabla_{\q}\phi(\w,\q)$. At each dual update, DRFA uniformly samples $U_s\subseteq[m]$ with $|U_s|=b$ and constructs
\[
\widehat{\mathbf R}_p(\w)=\frac{1}{p}\sum_{i\in U_s}\ell(\w;\z_i)\mathbf e_i.
\]
Since $\Pr(i\in U_s)=p$, we obtain $\E[\widehat{\mathbf R}_p(\w)]=\mathbf R(\w)$. Moreover, $G_\lambda^2:=\sup_{\w}\|\mathbf R(\w)\|_2^2\leq mG_R^2$, and, writing $\xi_i(\w;\z_i)=\ell(\w;\z_i)-R_i(\w)$,
\begin{equation*}
    \begin{split}
        V_p=\E\|\widehat{\mathbf R}_p(\w)-\mathbf R(\w)\|_2^2
        &=\left(\frac1p-1\right)\|\mathbf R(\w)\|_2^2+\frac1p\sum_{i=1}^m\E[\xi_i(\w;\z_i)^2]\\
        &\leq m\left[\left(\frac1p-1\right)G_R^2+\frac{\sigma_q^2}{p}\right].
    \end{split}
\end{equation*}
We then define $A_p:=G_\lambda^2+V_p=O\left(m/p\right)$.

Let $\w^{(t)}$ be the virtual averaged model and sample $t'$ uniformly from $\{s\tau+1,\ldots,(s+1)\tau\}$. Conditioning on the local trajectories, $\E[\tau\widehat{\mathbf R}_p(\w^{(t')})]=\sum_{t=s\tau+1}^{(s+1)\tau}\mathbf R(\w^{(t)})$, so the snapshot dual estimator remains unbiased. Applying the projection argument and the grouping argument of Lemmas~8--9 in~\citet{2020_DRFA}, the dual contribution is
\[
O\left(\gamma\tau A_p+\frac{D_\Lambda^2}{\gamma\tau\sqrt S}\right),\qquad S=\frac{T}{\tau}.
\]
The remaining primal terms introduce no polynomial dependence on $m$ or $p$. Indeed, for fixed $\q$, $\|\nabla_{\w}\phi(\w,\q)-\nabla_{\w}\phi(\w',\q)\|_2\leq L\|\w-\w'\|_2,$ the weighted gradient dissimilarity satisfies $\Gamma\leq4G^2$, and for $\widetilde{\w}\in\arg\min_{\u}\{\Phi(\u)+L\|\u-\w\|_2^2\}$, $\|\widetilde{\w}-\w\|_2\leq G/L.$
Thus, up to problem-dependent constants independent of $m$ and $p$, the proof of Theorem~2 yields
\begin{equation}\label{eqn:drfa_partial_master}
\frac1T\sum_{t=1}^T\E\|\nabla\Phi_{1/(2L)}(\w^{(t)})\|_2^2=O\left(\frac1{\eta T}+\eta\tau\sqrt S+\eta\tau+\eta+\gamma\tau A_p+\frac{D_\Lambda^2}{\gamma\tau\sqrt S}\right).
\end{equation}

\textbf{Local updates.} 
Set $\tau=T^{1/4}$, $S=T^{3/4}$, and $\eta=\Theta(T^{-3/4})$. Balancing the two dual terms in~(\ref{eqn:drfa_partial_master}) gives $\gamma=\Theta(A_p^{-1/2}T^{-7/16})$ and hence
\[
\frac1T\sum_{t=1}^T\E\|\nabla\Phi_{1/(2L)}(\w^{(t)})\|_2^2=O\left(T^{-1/8}+\sqrt{A_p}\,T^{-3/16}\right).
\]
For $r\sim\operatorname{Unif}[T]$, Jensen's inequality and $A_p=O(m/p)$ give
\[
\E\|\nabla\Phi_{1/(2L)}(\w^{(r)})\|_2=O\left(T^{-1/16}+\left(\frac{m}{p}\right)^{1/4}T^{-3/32}\right).
\]
Therefore,
\[
T=O\left(\varepsilon^{-16}+\left(\frac{m}{p}\right)^{8/3}\varepsilon^{-32/3}\right),
\]
Since each synchronization stage communicates $O(db)=O(dpm)$ entries, achieving an $\varepsilon$-stationary solution requires $O\left(\varepsilon^{-12}+m^2p^{-2}\varepsilon^{-8}\right)$ synchronization rounds and $O\left(dpm\varepsilon^{-12}+dm^3p^{-1}\varepsilon^{-8}\right)$ total communication.

\textbf{Synchronization at every step.} For $\tau=1$, we have $S=T$. Choosing $\eta=\Theta(T^{-3/4})$ and $\gamma=\Theta(A_p^{-1/2}T^{-1/4})$ gives
\[
\E\|\nabla\Phi_{1/(2L)}(\w^{(r)})\|_2=O\left(\left(\frac{m}{p}\right)^{1/4}T^{-1/8}\right).
\]
Therefore, $T=O\left(m^2p^{-2}\varepsilon^{-8}\right).$
Since every iteration is a synchronization stage and each stage communicates $O(db)=O(dpm)$ entries, achieving an $\varepsilon$-stationary solution requires $O\left(m^2p^{-2}\varepsilon^{-8}\right)$ synchronization rounds and $O\left(dm^3p^{-1}\varepsilon^{-8}\right)$ total communication.

In Table~\ref{tab:AFL-summary}, we report the dependence on $m$ and $\varepsilon$ for a fixed sampling fraction $p=\Theta(1)$ independent of both quantities. Absorbing the resulting $p$-dependent constants yields $O(\varepsilon^{-12}+m^2\varepsilon^{-8})$ synchronization rounds and $O(dm\varepsilon^{-12}+dm^3\varepsilon^{-8})$ communication for local-update DRFA, and $O(m^2\varepsilon^{-8})$ synchronization rounds and $O(dm^3\varepsilon^{-8})$ communication for $\tau=1$.

\section{Experimental Details}\label{app:experiments}
\subsection{Scaling with the Number of Workers}\label{app:exp_scaling}
In all experiments, we use a full-participation variant of \textsc{DRFA}~\citep{2020_DRFA} to compare optimization progress per synchronization stage under the same participation model as the other methods. We let every worker perform the local updates and use $\q^{(s)}$ to aggregate both the end-of-block models and the randomly selected snapshot models:
\[
\bar{\w}^{(s+1)}=\sum_{i=1}^m q_i^{(s)}\w_i^{((s+1)\tau)},\qquad \w^{(t')}=\sum_{i=1}^m q_i^{(s)}\w_i^{(t')}.
\]
The weighted aggregation preserves the weighting represented by the adaptive-sampling expectation. 
All methods are trained for $T=5000$ model updates, and for \textsc{AFL-BR} we set the block length to $B=\lceil\sqrt{T}\rceil=71$. The target worst-worker test accuracies for $m=20,50,100,500,$ and $1000$ are $0.72,0.71,0.70,0.48,$ and $0.46$, respectively.

\subsection{Communication Efficiency with Compression}\label{app:exp_compression}
For Fashion-MNIST~\citep{xiao2017fashionmnist} and CIFAR-10~\citep{CIFAR-10}, we construct label-skewed local datasets as follows. For each class $i\in[m]$, worker $i$ receives $80\%$ of the samples with label $i$, while the remaining $20\%$ are distributed uniformly among the other $m-1$ workers. Each worker's local data are then split into training and test sets using an $80/20$ split.
For the model architecture, we use a three-layer MLP with layer normalization and ReLU activations on Fashion-MNIST, and a two-block CNN followed by a linear classifier on CIFAR-10. Both models are trained using the cross-entropy loss.

Update rounds count every model update, whereas synchronization rounds count worker--server synchronization blocks. \textsc{FedAvg} and \textsc{DRFA} perform three local updates per synchronization, while \textsc{AFL}, \textsc{AFL-BR}, and both \textsc{AFL-Com} variants perform one update. Communication cost follows the accounting in Section~\ref{sec:exp:compression}, and the communication accounting for \textsc{DRFA} includes the additional model exchanges required by snapshotting.

\begin{figure*}[t]
\centering
\includegraphics[width=\textwidth]{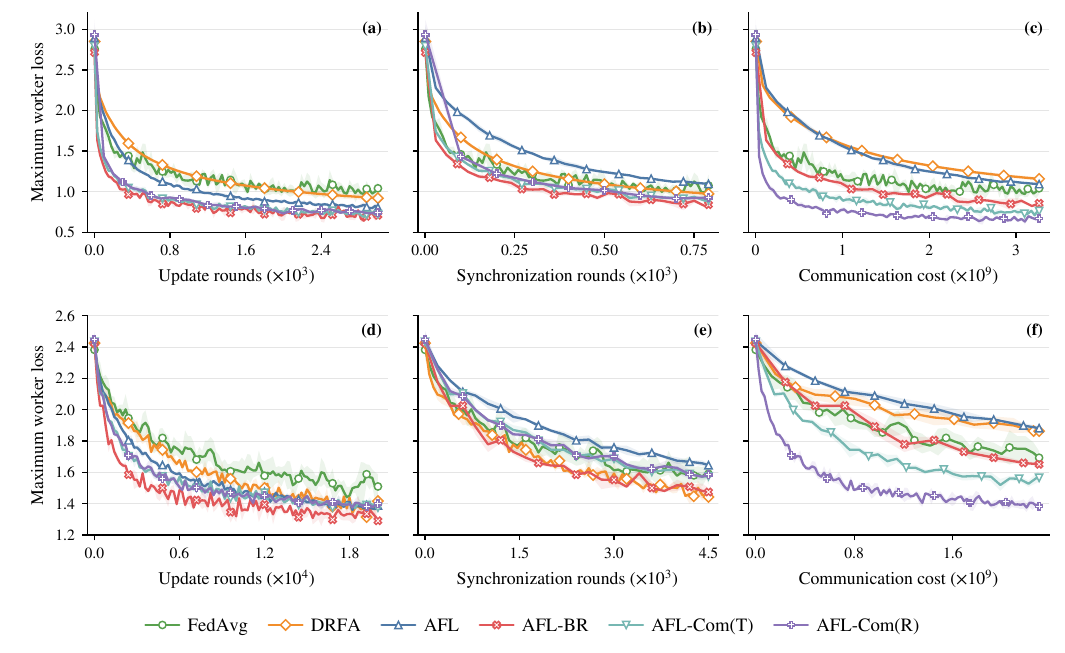}
\caption{\textbf{Maximum worker test loss.} The top and bottom rows correspond to Fashion-MNIST and CIFAR-10, respectively. From left to right, the columns report performance against update rounds, synchronization rounds, and cumulative communication cost.}
\label{fig:nonconvex_loss}
\end{figure*}

Figure~\ref{fig:nonconvex_loss} reports the maximum test loss across workers, corresponding to the robust objective evaluated on test data. Panels (a) and (d) show that \textsc{AFL-BR} generally achieves the lowest loss for a given number of updates, while both \textsc{AFL-Com} variants remain close despite compression.

Panels (b) and (e) show that \textsc{AFL-BR} remains competitive with \textsc{DRFA} in synchronization efficiency while using only one update per synchronization. At comparable communication budgets, panels (c) and (f) show that \textsc{AFL-Com}(R) achieves the lowest maximum loss on both datasets, while \textsc{AFL-Com}(T) also substantially outperforms the uncompressed AFL baselines. These results corroborate the accuracy-based communication-efficiency comparison in Section~\ref{sec:exp:compression}.

\section{Nonconvex Analysis} \label{app:nonconvex_lemmas}
\subsection{Common Lemmas for Nonconvex Analysis}\label{app:common_nonconvex_lemmas}
The following lemmas are shared by the analyses of \textsc{AFL-BR} and \textsc{AFL-Com}.

\begin{definition}  \label{def:weakly_convex}
    A function $f(\cdot)$ is called $L$-weakly convex if $f(\cdot)+\frac{L}{2}\|\cdot\|_2^2$ is convex.
\end{definition}

\begin{lemma}\label{lem:Phi_weakly_convex}\cite[Lemma 4.7]{2020_GDA_minimax}
	Under Assumption~\ref{ass:nonconvex_smooth}, for every fixed $\mathbf{q}\in\Delta_m$, the function $\phi(\cdot,\mathbf{q})$ is $L$-weakly convex. Moreover, $\Phi(\cdot)$ is $L$-weakly convex.
\end{lemma}

\begin{lemma}\label{lem:moreau_envelope_smooth}
Suppose $\Phi:\mathbb{R}^d\to\mathbb{R}$ is $L$-weakly convex and $G$-Lipschitz. Then $\Phi_{1/(2L)}$ is differentiable, and its gradient satisfies
\begin{equation*}\label{eqn:moreau_envelope_gradient_properties} \|\nabla\Phi_{1/(2L)}(\mathbf{x})-\nabla\Phi_{1/(2L)}(\mathbf{x}')\|_2\leq2L\|\mathbf{x}-\mathbf{x}'\|_2,\quad \|\nabla\Phi_{1/(2L)}(\mathbf{x})\|_2\leq G,\quad \forall\,\mathbf{x},\mathbf{x}'\in\mathbb{R}^d. 
\end{equation*}
\end{lemma}

\begin{lemma}[Perturbed Moreau-envelope descent]\label{lem:perturbed_moreau_descent}
Suppose Assumptions~\ref{ass:nonconvex_smooth} and~\ref{ass:nonconvex_lower} hold. Let $\u_t$, $\w_t$, and $\q_t\in\Delta_m$ be $\mathcal F_t$-measurable for every $t\in[T+1]$, and suppose that, for some $\eta_w>0$,
\begin{equation}\label{eqn:generic_virtual_update_new}
\u_{t+1}=\u_t-\eta_w\g_{w,t},\quad \mathbb E_t[\g_{w,t}]=\nabla_{\w}\phi(\w_t,\q_t),\quad \mathbb E_t\left[\left\|\g_{w,t}-\nabla_{\w}\phi(\w_t,\q_t)\right\|_2^2\right]\leq\sigma_w^2.
\end{equation}
Assume further that $\|\nabla_{\w}\phi(\w,\q)\|_2\leq G$ for every $\w\in\mathbb R^d$ and $\q\in\Delta_m$. Let $\mathcal J\subseteq[T]$ be a deterministic index set and define $\mathcal S=[T]\setminus\mathcal J$. For every $t\in[T]$, let $d_t=\|\u_t-\w_t\|_2$ and $\varepsilon_t^u=\Phi(\u_t)-\phi(\u_t,\q_t)$, and define $\Delta_u=\Phi_{1/(2L)}(\u_1)-\Phi_*$ and $\Gamma_w^2=G^2+\sigma_w^2$. Then
\begin{equation}\label{eqn:generic_moreau_descent_new}
\begin{aligned}
\frac{1}{T}\sum_{t=1}^T\mathbb E\left[\left\|\nabla\Phi_{1/(2L)}(\u_t)\right\|_2^2\right]
\leq{}&\frac{4\Delta_u}{\eta_wT}+\frac{8L}{T}\sum_{t\in\mathcal J}\mathbb E[\varepsilon_t^u]+\frac{4LG}{T}\sum_{t\in\mathcal J}\mathbb E[d_t]\\
&+4L\eta_w\Gamma_w^2+\frac{5\Gamma_w^2|\mathcal S|}{T}.
\end{aligned}
\end{equation}
\end{lemma}

\begin{lemma}[Blockwise dual-gap bound]\label{lem:blockwise_dual_gap}
Let $1\leq B\leq T$ and partition $[T]$ into $N=\lceil T/B\rceil$ consecutive blocks $\{\mathcal I_c\}_{c=1}^N$, where $\mathcal I_c=\{s_c,\ldots,e_c\}$ has length $B$ for every $c<N$ and length at most $B$ for $c=N$. Define the block-start and non-start index sets as $\mathcal S=\{s_1,\ldots,s_N\}$ and $\mathcal J=[T]\setminus\mathcal S$. Suppose that $\q_{s_c}=m^{-1}\mathbf1_m$ for every $c\in[N]$. 
For every $t\in\mathcal I_c$, define the pre-restart mirror-ascent iterate
\begin{equation}\label{eqn:blockwise_mirror_iterate_new}
\bar{\q}_{t+1}=\argmax_{\q\in\Delta_m}\left\{\eta_q\langle\widehat{\mathbf R}_t,\q\rangle-D_{\mathrm{KL}}(\q\|\q_t)\right\}.
\end{equation}
Assume that $\bar{\q}_{t+1}=\q_{t+1}$ for every $t<e_c$, while at the end of each nonfinal block the algorithm restarts and sets $\q_{e_c+1}=m^{-1}\mathbf1_m$. Suppose Assumption~\ref{ass:nonconvex_dual_oracle} holds, the round-$t$ oracle $\widehat{\mathbf R}_t$ is evaluated at $\w_t$, and every $R_i$ is $G$-Lipschitz. Define $\varepsilon_t^q=\Phi(\w_t)-\phi(\w_t,\q_t)$. Then
\begin{equation}\label{eqn:blockwise_dual_gap_general_new}
\frac{1}{T}\sum_{t\in\mathcal J}\mathbb E[\varepsilon_t^q]\leq\frac{2G(B+2)}{T}\sum_{t=1}^{T-1}\mathbb E[\|\w_{t+1}-\w_t\|_2]+\frac{2\ln m}{\eta_qB}+\frac{\eta_q\sigma_q^2}{2}.
\end{equation}
\end{lemma}

\begin{lemma}\label{lem:compression_residuals}
Under Assumption~\ref{ass:nonconvex_primal_oracle}, consider Algorithm~\ref{alg:aflcom_br} and define $\e_t:=\sum_{i=1}^m\e_t^{(i)}$, $\g_{w,t}:=\sum_{i=1}^m q_{t,i}\g_{w,t}^{(i)}$, and $\rb_t:=\e_t+\hateb_t$. For every $t\leq T+1$, a general $\delta$-approximate compressor satisfies
\begin{equation*}\label{eqn:compression_residual_uniform_bounds}
\mathbb E\left[\|\e_t\|_2\right]\leq\frac{2-\delta}{\delta}\Gamma_w,\qquad \mathbb E\left[\|\hateb_t\|_2\right]\leq\frac{8}{\delta^2}\Gamma_w.
\end{equation*}
Moreover, since $\Delta_{w,t}=\g_{w,t}+\rb_t-\rb_{t+1}$,
\begin{equation}\label{eqn:compression_residual_average_bounds}
\frac{1}{T}\sum_{t=1}^T\mathbb E\left[\|\rb_t\|_2\right]\leq\frac{8+2\delta-\delta^2}{\delta^2}\Gamma_w,\qquad \frac{1}{T}\sum_{t=1}^T\mathbb E\left[\|\Delta_{w,t}\|_2\right]\leq\frac{16+4\delta-\delta^2}{\delta^2}\Gamma_w.
\end{equation}
If $\mathcal C\in\mathcal C_{\mathrm{AI}}$ and Algorithm~\ref{alg:aflcom_br} uses SR, then $\hateb_t=\mathbf0_d$ pathwise for every $t\leq T+1$, and the bounds~(\ref{eqn:compression_residual_average_bounds}) improve to
\begin{equation*}\label{eqn:compression_ai_average_bounds}
\frac{1}{T}\sum_{t=1}^T\mathbb E\left[\|\rb_t\|_2\right]\leq\frac{2\Gamma_w}{\delta},\qquad \frac{1}{T}\sum_{t=1}^T\mathbb E\left[\|\Delta_{w,t}\|_2\right]\leq\frac{4\Gamma_w}{\delta}.
\end{equation*}
\end{lemma}

\subsection{Proof of Theorem~\ref{thm:uafl_br}}\label{app:proof_uafl_br}
Let $\mathcal S$ be the set of block-start indices and let $\mathcal J=[T]\setminus\mathcal S$. For Algorithm~\ref{alg:uafl_br}, set $\u_t=\w_t$. The aggregated stochastic gradient is $\g_{w,t}=\sum_{i=1}^m q_{t,i}\g_{w,t}^{(i)}$. By Assumption~\ref{ass:nonconvex_primal_oracle},
\begin{equation}\label{eqn:uafl_aggregate_unbiased_new}
\mathbb E_t[\g_{w,t}]=\sum_{i=1}^m q_{t,i}\nabla R_i(\w_t)=\nabla_{\w}\phi(\w_t,\q_t).
\end{equation}
Let $\boldsymbol\zeta_{t,i}=\g_{w,t}^{(i)}-\nabla R_i(\w_t)$. Since $\|\sum_iq_{t,i}\boldsymbol\zeta_{t,i}\|_2^2\leq\sum_iq_{t,i}\|\boldsymbol\zeta_{t,i}\|_2^2$ pointwise,
\begin{equation}\label{eqn:uafl_aggregate_variance_new}
\mathbb E_t\left[\left\|\g_{w,t}-\nabla_{\w}\phi(\w_t,\q_t)\right\|_2^2\right]\leq\sum_{i=1}^m q_{t,i}\mathbb E_t[\|\boldsymbol\zeta_{t,i}\|_2^2]\leq\sigma_w^2.
\end{equation}
Moreover, $\|\nabla_{\w}\phi(\w_t,\q_t)\|_2\leq\sum_iq_{t,i}\|\nabla R_i(\w_t)\|_2\leq G$. Hence, $\mathbb E_t[\|\g_{w,t}\|_2^2]\leq\Gamma_w^2$ and $\mathbb E_t[\|\g_{w,t}\|_2]\leq\Gamma_w$.

Since $\u_t=\w_t$, we have $d_t=0$, $\varepsilon_t^u=\varepsilon_t^q$, and $\Delta_u=\Delta_\Phi$. Applying Lemma~\ref{lem:perturbed_moreau_descent} gives
\begin{equation}\label{eqn:uafl_moreau_pre_gap_new}
\frac1T\sum_{t=1}^T\mathbb E\left[\left\|\nabla\Phi_{1/(2L)}(\w_t)\right\|_2^2\right]\leq\frac{4\Delta_\Phi}{\eta_wT}+\frac{8L}{T}\sum_{t\in\mathcal J}\mathbb E[\varepsilon_t^q]+4L\eta_w\Gamma_w^2+\frac{5\Gamma_w^2|\mathcal S|}{T}.
\end{equation}
The primal update satisfies $\w_{t+1}-\w_t=-\eta_w\g_{w,t}$, and therefore
\begin{equation}\label{eqn:uafl_movement_new}
\sum_{t=1}^{T-1}\mathbb E[\|\w_{t+1}-\w_t\|_2]\leq\eta_w\Gamma_wT.
\end{equation}
Applying Lemma~\ref{lem:blockwise_dual_gap} with $\widehat{\mathbf R}_t=\g_{q,t}$ and then~(\ref{eqn:uafl_movement_new}) gives
\begin{equation}\label{eqn:uafl_dual_gap_new}
\frac1T\sum_{t\in\mathcal J}\mathbb E[\varepsilon_t^q]\leq2G\Gamma_w\eta_w(B+2)+\frac{2\ln m}{\eta_qB}+\frac{\eta_q\sigma_q^2}{2}.
\end{equation}
Substituting~(\ref{eqn:uafl_dual_gap_new}) into~(\ref{eqn:uafl_moreau_pre_gap_new}) and using $|\mathcal S|=N\leq2T/B$ yields
\begin{equation}\label{eqn:uafl_master_new}
\begin{aligned}
\frac1T\sum_{t=1}^T\mathbb E\left[\left\|\nabla\Phi_{1/(2L)}(\w_t)\right\|_2^2\right]
\leq{}&\frac{4\Delta_\Phi}{\eta_wT}+16LG\Gamma_w\eta_w(B+2)+\frac{16L\ln m}{\eta_qB}+4L\eta_q\sigma_q^2\\
&+4L\eta_w\Gamma_w^2+\frac{10\Gamma_w^2}{B}.
\end{aligned}
\end{equation}
Because $T\geq9$ and $B=\lceil\sqrt T\rceil$, we have $B\geq\sqrt T$ and $B+2\leq2\sqrt T$. Substituting $\eta_w=(8LT^{3/4})^{-1}$ gives
\begin{equation}\label{eqn:uafl_primal_terms_new}
\frac{4\Delta_\Phi}{\eta_wT}=\frac{32L\Delta_\Phi}{T^{1/4}},
\qquad
16LG\Gamma_w\eta_w(B+2)\leq\frac{4G\Gamma_w}{T^{1/4}},
\qquad
4L\eta_w\Gamma_w^2=\frac{\Gamma_w^2}{2T^{3/4}}.
\end{equation}
Substituting $\eta_q=2\sqrt{\ln m}/(\sigma_qT^{1/4})$ and using $B\geq\sqrt T$ gives
\begin{equation}\label{eqn:uafl_dual_terms_new}
\frac{16L\ln m}{\eta_qB}+4L\eta_q\sigma_q^2\leq\frac{8L\sigma_q\sqrt{\ln m}}{T^{1/4}}+\frac{8L\sigma_q\sqrt{\ln m}}{T^{1/4}}=\frac{16L\sigma_q\sqrt{\ln m}}{T^{1/4}}.
\end{equation}
Finally, $10\Gamma_w^2/B\leq10\Gamma_w^2/T^{1/2}$. Combining~(\ref{eqn:uafl_master_new})--(\ref{eqn:uafl_dual_terms_new}) yields
\begin{equation}\label{eqn:uafl_average_rate_new}
\frac1T\sum_{t=1}^T\mathbb E\left[\left\|\nabla\Phi_{1/(2L)}(\w_t)\right\|_2^2\right]
\leq \frac{32L\Delta_\Phi}{T^{1/4}}+\frac{4G\Gamma_w}{T^{1/4}}+\frac{16L\sigma_q\sqrt{\ln m}}{T^{1/4}}+\frac{10\Gamma_w^2}{T^{1/2}}+\frac{\Gamma_w^2}{2T^{3/4}}.
\end{equation}
Since $r$ is sampled uniformly and independently from $[T]$, the left-hand side of~(\ref{eqn:uafl_average_rate_new}) equals $\mathbb E[\|\nabla\Phi_{1/(2L)}(\w_r)\|_2^2]$. This completes the proof.

\subsection{Proofs of Theorems~\ref{thm:aflcom_br_general} and~\ref{thm:aflcom_br_ai}}\label{app:proof_aflcom_br}
We first derive a master inequality common to both compressor classes.

Let $\mathcal S$ be the set of block-start indices and let $\mathcal J=[T]\setminus\mathcal S$. Define the aggregated uplink residual $\e_t=\sum_{i=1}^m\e_t^{(i)}$ and the total residual $\rb_t=\e_t+\hateb_t$. The uplink EF recursion gives $\hatgb_{w,t}=\g_{w,t}+\e_t-\e_{t+1}$. Combining this identity with the downlink EF recursion yields
\begin{equation}\label{eqn:aflcom_total_residual_identity_new}
\Delta_{w,t}=\hatgb_{w,t}+\hateb_t-\hateb_{t+1}=\g_{w,t}+\rb_t-\rb_{t+1}.
\end{equation}
Introduce the virtual iterate $\u_t=\w_t-\eta_w\rb_t.$ Using $\w_{t+1}=\w_t-\eta_w\Delta_{w,t}$ and~(\ref{eqn:aflcom_total_residual_identity_new}), we obtain
\begin{equation}\label{eqn:aflcom_virtual_recursion_new}
\u_{t+1}=\w_{t+1}-\eta_w\rb_{t+1}=\w_t-\eta_w\Delta_{w,t}-\eta_w\rb_{t+1}=\u_t-\eta_w\g_{w,t}.
\end{equation}
The same argument as in~(\ref{eqn:uafl_aggregate_unbiased_new})--(\ref{eqn:uafl_aggregate_variance_new}) gives
\begin{equation*}
\mathbb E_t[\g_{w,t}]=\nabla_{\w}\phi(\w_t,\q_t),
\qquad
\mathbb E_t\left[\left\|\g_{w,t}-\nabla_{\w}\phi(\w_t,\q_t)\right\|_2^2\right]\leq\sigma_w^2,
\qquad
\mathbb E_t[\|\g_{w,t}\|_2]\leq\Gamma_w.
\end{equation*}
All residuals are initialized at zero, so $\u_1=\w_1$ and $\Delta_u=\Delta_\Phi$. Moreover, $d_t=\|\u_t-\w_t\|_2=\eta_w\|\rb_t\|_2.$
Applying Lemma~\ref{lem:perturbed_moreau_descent} to~(\ref{eqn:aflcom_virtual_recursion_new}) gives
\begin{equation}\label{eqn:aflcom_virtual_pre_gap_new}
\begin{aligned}
\frac1T\sum_{t=1}^T\mathbb E\left[\left\|\nabla\Phi_{1/(2L)}(\u_t)\right\|_2^2\right]
\leq{}&\frac{4\Delta_\Phi}{\eta_wT}+\frac{8L}{T}\sum_{t\in\mathcal J}\mathbb E[\varepsilon_t^u]+\frac{4LG\eta_w}{T}\sum_{t\in\mathcal J}\mathbb E[\|\rb_t\|_2]\\
&+4L\eta_w\Gamma_w^2+\frac{5\Gamma_w^2|\mathcal S|}{T}.
\end{aligned}
\end{equation}
Because $\Phi$ and $\phi(\cdot,\q_t)$ are both $G$-Lipschitz,
\begin{equation}\label{eqn:aflcom_gap_transfer_new}
\varepsilon_t^u=\Phi(\u_t)-\phi(\u_t,\q_t)\leq\Phi(\w_t)-\phi(\w_t,\q_t)+2G\|\u_t-\w_t\|_2=\varepsilon_t^q+2G\eta_w\|\rb_t\|_2.
\end{equation}
Substituting~(\ref{eqn:aflcom_gap_transfer_new}) into~(\ref{eqn:aflcom_virtual_pre_gap_new}), enlarging the residual sum from $\mathcal J$ to $[T]$, and using $|\mathcal S|\leq2T/B$ yield
\begin{equation}\label{eqn:aflcom_virtual_with_gap_new}
\begin{aligned}
\frac1T\sum_{t=1}^T\mathbb E\left[\left\|\nabla\Phi_{1/(2L)}(\u_t)\right\|_2^2\right]
\leq{}&\frac{4\Delta_\Phi}{\eta_wT}+\frac{8L}{T}\sum_{t\in\mathcal J}\mathbb E[\varepsilon_t^q]+\frac{20LG\eta_w}{T}\sum_{t=1}^T\mathbb E[\|\rb_t\|_2]\\
&+4L\eta_w\Gamma_w^2+\frac{10\Gamma_w^2}{B}.
\end{aligned}
\end{equation}
The actual update satisfies $\|\w_{t+1}-\w_t\|_2=\eta_w\|\Delta_{w,t}\|_2$. Applying Lemma~\ref{lem:blockwise_dual_gap} gives
\begin{equation}\label{eqn:aflcom_dual_gap_new}
\frac1T\sum_{t\in\mathcal J}\mathbb E[\varepsilon_t^q]\leq\frac{2G\eta_w(B+2)}{T}\sum_{t=1}^T\mathbb E[\|\Delta_{w,t}\|_2]+\frac{2\ln m}{\eta_qB}+\frac{\eta_q\sigma_q^2}{2}.
\end{equation}
Combining~(\ref{eqn:aflcom_virtual_with_gap_new}) and~(\ref{eqn:aflcom_dual_gap_new}) yields
\begin{equation}\label{eqn:aflcom_virtual_master_new}
\begin{aligned}
\frac1T\sum_{t=1}^T\mathbb E\left[\left\|\nabla\Phi_{1/(2L)}(\u_t)\right\|_2^2\right]
\leq &\frac{4\Delta_\Phi}{\eta_wT}+\frac{16LG\eta_w(B+2)}{T}\sum_{t=1}^T\mathbb E[\|\Delta_{w,t}\|_2]+\frac{16L\ln m}{\eta_qB}\\
&+4L\eta_q\sigma_q^2+\frac{20LG\eta_w}{T}\sum_{t=1}^T\mathbb E[\|\rb_t\|_2]+4L\eta_w\Gamma_w^2+\frac{10\Gamma_w^2}{B}.
\end{aligned}
\end{equation}
Let $\G_t^w=\nabla\Phi_{1/(2L)}(\w_t)$ and $\G_t^u=\nabla\Phi_{1/(2L)}(\u_t)$. Since $\nabla\Phi_{1/(2L)}$ is $2L$-Lipschitz and both gradients have norm at most $G$ by Lemma~\ref{lem:moreau_envelope_smooth}, we have
\begin{equation}\label{eqn:aflcom_stationarity_transfer_new}
\|\G_t^w\|_2^2-\|\G_t^u\|_2^2\leq\|\G_t^w-\G_t^u\|_2(\|\G_t^w\|_2+\|\G_t^u\|_2)\leq4LG\|\w_t-\u_t\|_2=4LG\eta_w\|\rb_t\|_2.
\end{equation}
Taking expectations, averaging~(\ref{eqn:aflcom_stationarity_transfer_new}), and applying~(\ref{eqn:aflcom_virtual_master_new}) give
\begin{equation}\label{eqn:aflcom_actual_master_new}
\begin{aligned}
\frac1T\sum_{t=1}^T\mathbb E\left[\left\|\nabla\Phi_{1/(2L)}(\w_t)\right\|_2^2\right]
\leq{}&\frac{4\Delta_\Phi}{\eta_wT}+\frac{16LG\eta_w(B+2)}{T}\sum_{t=1}^T\mathbb E[\|\Delta_{w,t}\|_2]+\frac{16L\ln m}{\eta_qB}\\
&+4L\eta_q\sigma_q^2+\frac{24LG\eta_w}{T}\sum_{t=1}^T\mathbb E[\|\rb_t\|_2]+4L\eta_w\Gamma_w^2+\frac{10\Gamma_w^2}{B}.
\end{aligned}
\end{equation}
Since $r$ is sampled uniformly from $[T]$ independently of the algorithmic randomness,
\begin{equation}\label{eqn:aflcom_random_output_new}
\mathbb E\left[\left\|\nabla\Phi_{1/(2L)}(\w_r)\right\|_2^2\right]=\frac{1}{T}\sum_{t=1}^T\mathbb E\left[\left\|\nabla\Phi_{1/(2L)}(\w_t)\right\|_2^2\right].
\end{equation}

\paragraph{Proof of Theorem~\ref{thm:aflcom_br_general}.} For a general $\delta$-approximate compressor, Lemma~\ref{lem:compression_residuals} gives
\begin{equation*}
\frac{1}{T}\sum_{t=1}^T\mathbb E\left[\|\rb_t\|_2\right]\leq\frac{8+2\delta-\delta^2}{\delta^2}\Gamma_w,\qquad \frac{1}{T}\sum_{t=1}^T\mathbb E\left[\|\Delta_{w,t}\|_2\right]\leq\frac{16+4\delta-\delta^2}{\delta^2}\Gamma_w.
\end{equation*}
Substituting these bounds into~(\ref{eqn:aflcom_actual_master_new}) yields
\begin{equation*}
\begin{aligned}
\frac{1}{T}\sum_{t=1}^T\mathbb E\left[\left\|\nabla\Phi_{1/(2L)}(\w_t)\right\|_2^2\right]
\leq{}&\frac{4\Delta_\Phi}{\eta_wT}+\frac{16LG\Gamma_w\eta_w(B+2)(16+4\delta-\delta^2)}{\delta^2}+\frac{16L\ln m}{\eta_qB}\\
&+4L\eta_q\sigma_q^2+\frac{24LG\Gamma_w\eta_w(8+2\delta-\delta^2)}{\delta^2}+4L\eta_w\Gamma_w^2+\frac{10\Gamma_w^2}{B}.
\end{aligned}
\end{equation*}
Using $B=\lceil\sqrt T\rceil$, $B\geq\sqrt T$, $B+2\leq2\sqrt T$, $\eta_w=(8LT^{3/4})^{-1}$, and $\eta_q=2\sqrt{\ln m}/(\sigma_qT^{1/4})$ gives
\begin{equation}\label{eqn:aflcom_general_average_rate_new}
\begin{aligned}
\frac{1}{T}\sum_{t=1}^T\mathbb E\left[\left\|\nabla\Phi_{1/(2L)}(\w_t)\right\|_2^2\right]
\leq{}&\frac{32L\Delta_\Phi}{T^{1/4}}+\frac{4G\Gamma_w(16+4\delta-\delta^2)}{\delta^2T^{1/4}}+\frac{16L\sigma_q\sqrt{\ln m}}{T^{1/4}}+\frac{10\Gamma_w^2}{T^{1/2}}\\
&+\frac{3G\Gamma_w(8+2\delta-\delta^2)}{\delta^2T^{3/4}}+\frac{\Gamma_w^2}{2T^{3/4}}.
\end{aligned}
\end{equation}
Combining~(\ref{eqn:aflcom_random_output_new}) and~(\ref{eqn:aflcom_general_average_rate_new}) proves~(\ref{eqn:aflcom_br_general_rate_new}).

\paragraph{Proof of Theorem~\ref{thm:aflcom_br_ai}.} 
If $\mathcal C\in\mathcal C_{\mathrm{AI}}$ and SR is used, Lemma~\ref{lem:compression_residuals} gives
\begin{equation*}
\frac{1}{T}\sum_{t=1}^T\mathbb E\left[\|\rb_t\|_2\right]\leq\frac{2\Gamma_w}{\delta},\qquad \frac{1}{T}\sum_{t=1}^T\mathbb E\left[\|\Delta_{w,t}\|_2\right]\leq\frac{4\Gamma_w}{\delta}.
\end{equation*}
Substituting these bounds into~(\ref{eqn:aflcom_actual_master_new}) gives
\begin{equation*}\label{eqn:aflcom_ai_master_new}
\begin{aligned}
\frac{1}{T}\sum_{t=1}^T\mathbb E\left[\left\|\nabla\Phi_{1/(2L)}(\w_t)\right\|_2^2\right]
\leq{}&\frac{4\Delta_\Phi}{\eta_wT}+\frac{64LG\Gamma_w\eta_w(B+2)}{\delta}+\frac{16L\ln m}{\eta_qB}+4L\eta_q\sigma_q^2\\
&+\frac{48LG\Gamma_w\eta_w}{\delta}+4L\eta_w\Gamma_w^2+\frac{10\Gamma_w^2}{B}.
\end{aligned}
\end{equation*}
Using the same parameter choices yields
\begin{equation}\label{eqn:aflcom_ai_average_rate_new}
\begin{aligned}
\frac{1}{T}\sum_{t=1}^T\mathbb E\left[\left\|\nabla\Phi_{1/(2L)}(\w_t)\right\|_2^2\right]
\leq{}&\frac{32L\Delta_\Phi}{T^{1/4}}+\frac{16G\Gamma_w}{\delta T^{1/4}}+\frac{16L\sigma_q\sqrt{\ln m}}{T^{1/4}}+\frac{10\Gamma_w^2}{T^{1/2}}\\
&+\frac{6G\Gamma_w}{\delta T^{3/4}}+\frac{\Gamma_w^2}{2T^{3/4}}.
\end{aligned}
\end{equation}
Combining~(\ref{eqn:aflcom_random_output_new}) and~(\ref{eqn:aflcom_ai_average_rate_new}) proves Theorem~\ref{thm:aflcom_br_ai}.

\section{Proof of Lemmas}
\subsection{Proof of Lemma~\ref{lem:moreau_parameter_comparison}}
Because $f$ is $\rho$-weakly convex, the function $h(\y):=f(\y)+\frac{\rho}{2}\|\y\|_2^2$ is convex. For any $\lambda<1/\rho$, the proximal objective can be written as
\[
f(\y)+\frac{1}{2\lambda}\|\y-\x\|_2^2=h(\y)+\frac{1}{2}\left(\frac{1}{\lambda}-\rho\right)\|\y\|_2^2-\frac{1}{\lambda}\langle\x,\y\rangle+\frac{1}{2\lambda}\|\x\|_2^2.
\]
Since $1/\lambda-\rho>0$, this objective is $(1/\lambda-\rho)$-strongly convex and therefore has a unique minimizer. The same argument applies to $\mu<1/\rho$. Define
\[
\y_\lambda=\argmin_{\y\in\mathbb R^d}\left\{f(\y)+\frac{1}{2\lambda}\|\y-\x\|_2^2\right\},\qquad \y_\mu=\argmin_{\y\in\mathbb R^d}\left\{f(\y)+\frac{1}{2\mu}\|\y-\x\|_2^2\right\}.
\]
The uniqueness of these minimizers and Danskin's theorem imply that the corresponding Moreau envelopes are differentiable at $\x$, with
\[
\g_\lambda:=\nabla f_\lambda(\x)=\frac{\x-\y_\lambda}{\lambda},\qquad \g_\mu:=\nabla f_\mu(\x)=\frac{\x-\y_\mu}{\mu}.
\]
The first-order optimality conditions for the proximal subproblems give
\[
\mathbf 0\in\partial f(\y_\lambda)+\frac{1}{\lambda}(\y_\lambda-\x),\qquad \mathbf 0\in\partial f(\y_\mu)+\frac{1}{\mu}(\y_\mu-\x).
\]
Hence, $\g_\lambda\in\partial f(\y_\lambda)$, and $\g_\mu\in\partial f(\y_\mu)$.
Since $f$ is $\rho$-weakly convex, its subdifferential is $\rho$-hypomonotone. Therefore,
\[
\langle\g_\lambda-\g_\mu,\y_\lambda-\y_\mu\rangle\geq-\rho\|\y_\lambda-\y_\mu\|_2^2.
\]
Substituting $\y_\lambda=\x-\lambda\g_\lambda$ and $\y_\mu=\x-\mu\g_\mu$ into the hypomonotonicity inequality, its left-hand side becomes
\[
\begin{aligned}
\langle\g_\lambda-\g_\mu,-\lambda\g_\lambda+\mu\g_\mu\rangle
&=-\lambda\|\g_\lambda\|_2^2-\mu\|\g_\mu\|_2^2+(\lambda+\mu)\langle\g_\lambda,\g_\mu\rangle,
\end{aligned}
\]
whereas the squared norm on the right-hand side satisfies
\[
\|-\lambda\g_\lambda+\mu\g_\mu\|_2^2=\lambda^2\|\g_\lambda\|_2^2+\mu^2\|\g_\mu\|_2^2-2\lambda\mu\langle\g_\lambda,\g_\mu\rangle.
\]
Combining these two expansions and rearranging gives
\begin{equation} \label{eqn:dis:SGDA:2}
\lambda(1-\rho\lambda)\|\g_\lambda\|_2^2+\mu(1-\rho\mu)\|\g_\mu\|_2^2\leq(\lambda+\mu-2\rho\lambda\mu)\langle\g_\lambda,\g_\mu\rangle \leq(\lambda+\mu-2\rho\lambda\mu)\|\g_\lambda\|_2\|\g_\mu\|_2.
\end{equation}

We first consider the case $\g_\mu=\mathbf 0$. The preceding inequality~(\ref{eqn:dis:SGDA:2}) then reduces to
\[
\lambda(1-\rho\lambda)\|\g_\lambda\|_2^2\leq0.
\]
Since $\lambda>0$ and $1-\rho\lambda>0$, it follows that $\g_\lambda=\mathbf 0$, and the desired inequality holds.

Now suppose that $\g_\mu\neq\mathbf 0$, and define $r:=\|\g_\lambda\|_2/\|\g_\mu\|_2$.
Dividing both sides of~(\ref{eqn:dis:SGDA:2}) by $\|\g_\mu\|_2^2$ yields
\[
\lambda(1-\rho\lambda)r^2-(\lambda+\mu-2\rho\lambda\mu)r+\mu(1-\rho\mu)\leq0.
\]
The quadratic polynomial on the left factors as
\[
\lambda(1-\rho\lambda)\left(r-\frac{\mu}{\lambda}\right)\left(r-\frac{1-\rho\mu}{1-\rho\lambda}\right).
\]
Because $\lambda<1/\rho$, its leading coefficient $\lambda(1-\rho\lambda)$ is positive. Moreover, $\mu\leq\lambda$ implies
\[
\frac{1-\rho\mu}{1-\rho\lambda}-\frac{\mu}{\lambda}=\frac{\lambda-\mu}{\lambda(1-\rho\lambda)}\geq0,
\]
so $(1-\rho\mu)/(1-\rho\lambda)$ is the larger of the two roots. Since the quadratic is nonpositive, $r$ lies between its two roots, and in particular,
\[
r\leq\frac{1-\rho\mu}{1-\rho\lambda}.
\]
Recalling the definition of $r$, we conclude that
\[
\|\nabla f_\lambda(\x)\|_2=\|\g_\lambda\|_2\leq\frac{1-\rho\mu}{1-\rho\lambda}\|\g_\mu\|_2=\frac{1-\rho\mu}{1-\rho\lambda}\|\nabla f_\mu(\x)\|_2.
\]

\subsection{Proof of Lemma~\ref{lem:Phi_weakly_convex}}
By Assumption~\ref{ass:nonconvex_smooth}, each $R_i$ is $L$-smooth. Hence,
for any $\mathbf{x},\mathbf{y}\in\mathbb{R}^d$,
\[
	R_i(\mathbf{y})	
    \ge R_i(\mathbf{x})+\langle \nabla R_i(\mathbf{x}),\mathbf{y}-\mathbf{x}\rangle-\frac{L}{2}\|\mathbf{y}-\mathbf{x}\|_2^2 .
\]
Define $f_i(\mathbf{x})=R_i(\mathbf{x})+\frac{L}{2}\|\mathbf{x}\|_2^2$. Then $\nabla f_i(\mathbf{x})=\nabla R_i(\mathbf{x})+L\mathbf{x}.$
Using the above lower bound, we have
\[
\begin{split}
	f_i(\mathbf{y})
	&= R_i(\mathbf{y})+\frac{L}{2}\|\mathbf{y}\|_2^2 \\
	&\ge
	R_i(\mathbf{x})+\langle \nabla R_i(\mathbf{x}),\mathbf{y}-\mathbf{x}\rangle-\frac{L}{2}\|\mathbf{y}-\mathbf{x}\|_2^2+\frac{L}{2}\|\mathbf{y}\|_2^2 \\
	&=R_i(\mathbf{x})+\frac{L}{2}\|\mathbf{x}\|_2^2+\left\langle\nabla R_i(\mathbf{x})+L\mathbf{x},\mathbf{y}-\mathbf{x}\right\rangle \\
	&=	f_i(\mathbf{x})+\langle \nabla f_i(\mathbf{x}),\mathbf{y}-\mathbf{x}\rangle .
\end{split}
\]
Therefore $f_i$ is convex, and hence $R_i$ is $L$-weakly convex. 

For any fixed $\mathbf{q}\in\Delta_m$, since $q_i\ge 0$ and
$\sum_{i=1}^m q_i=1$, we have
\[
\phi(\cdot,\mathbf{q})+\frac{L}{2}\|\cdot\|_2^2=\sum_{i=1}^m q_i R_i(\cdot)	+\frac{L}{2}\|\cdot\|_2^2 = \sum_{i=1}^m q_i\left(	R_i(\cdot)+\frac{L}{2}\|\cdot\|_2^2
\right).
\]
The right-hand side is a convex combination of convex functions, and is
therefore convex. Hence $\phi(\cdot,\mathbf{q})$ is $L$-weakly convex for
every $\mathbf{q}\in\Delta_m$.

Moreover,
\[
\Phi(\cdot)+\frac{L}{2}\|\cdot\|_2^2
=\max_{\mathbf{q}\in\Delta_m}\phi(\cdot,\mathbf{q})+	\frac{L}{2}\|\cdot\|_2^2 = \max_{\mathbf{q}\in\Delta_m} \left\{\phi(\cdot,\mathbf{q})+\frac{L}{2}\|\cdot\|_2^2\right\}.
\]
Since the pointwise maximum of convex functions is convex,
$\Phi(\cdot)+\frac{L}{2}\|\cdot\|_2^2$ is convex. Therefore, $\Phi$ is
$L$-weakly convex.

\subsection{Proof of Lemma~\ref{lem:moreau_envelope_smooth}}
For any $\mathbf{x}\in\mathbb{R}^d$, define
\begin{equation}\label{eqn:moreau_prox_point} \mathbf{y}(\mathbf{x})=\argmin_{\mathbf{y}\in\mathbb{R}^d}\left\{\Phi(\mathbf{y})+L\|\mathbf{y}-\mathbf{x}\|_2^2\right\}. \end{equation}
Since $\Phi$ is $L$-weakly convex, the function $\mathbf{y}\mapsto\Phi(\mathbf{y})+\frac{L}{2}\|\mathbf{y}\|_2^2$ is convex. Hence, for every fixed $\mathbf{x}$, the objective in~(\ref{eqn:moreau_prox_point}) is strongly convex, and $\mathbf{y}(\mathbf{x})$ is uniquely defined.

The first-order optimality condition for~(\ref{eqn:moreau_prox_point}) gives $\mathbf{0}\in\partial\Phi(\mathbf{y}(\mathbf{x}))+2L(\mathbf{y}(\mathbf{x})-\mathbf{x}).$
Then, we define $\mathbf{G}(\mathbf{x})=2L(\mathbf{x}-\mathbf{y}(\mathbf{x}))\in\partial \Phi(\mathbf{y}(\mathbf{x}))$. 
Since the minimizer in~(\ref{eqn:moreau_prox_point}) is unique, Danskin's theorem implies that the Moreau envelope is differentiable and
\begin{equation}\label{eqn:moreau_gradient_formula} \nabla\Phi_{1/(2L)}(\mathbf{x})=2L(\mathbf{x}-\mathbf{y}(\mathbf{x}))=\mathbf{G}(\mathbf{x}). \end{equation}

We first establish the uniform bound on the Moreau-envelope gradient. Since $\Phi$ is $G$-Lipschitz, every subgradient $\mathbf{v}\in\partial\Phi(\mathbf{y})$ satisfies $\|\mathbf{v}\|_2\leq G$. Combining this property with~$\mathbf{G}(\mathbf{x})\in\partial \Phi(\mathbf{y}(\mathbf{x}))$ and~(\ref{eqn:moreau_gradient_formula}) yields
\begin{equation}\label{eqn:moreau_gradient_norm_bound} \|\nabla\Phi_{1/(2L)}(\mathbf{x})\|_2=\|\mathbf{G}(\mathbf{x})\|_2\leq G. \end{equation}

It remains to prove that $\mathbf{G}(\cdot)$ is $2L$-Lipschitz. For any $\mathbf{x},\mathbf{x}'\in\mathbb{R}^d$, the optimality condition gives
$\mathbf{G}(\mathbf{x})\in\partial\Phi(\mathbf{y}(\mathbf{x}))$ and $\mathbf{G}(\mathbf{x}')\in\partial\Phi(\mathbf{y}(\mathbf{x}'))$.
Since $\Phi$ is $L$-weakly convex, its subdifferential is $L$-hypomonotone. Therefore,
\begin{equation}\label{eqn:moreau_hypomonotonicity} \left\langle\mathbf{G}(\mathbf{x})-\mathbf{G}(\mathbf{x}'),\mathbf{y}(\mathbf{x})-\mathbf{y}(\mathbf{x}')\right\rangle\geq-L\|\mathbf{y}(\mathbf{x})-\mathbf{y}(\mathbf{x}')\|_2^2. \end{equation}
By the definition of $\mathbf{G}$,
\begin{equation}\label{eqn:moreau_gradient_difference} \mathbf{G}(\mathbf{x})-\mathbf{G}(\mathbf{x}')=2L\left(\mathbf{x}-\mathbf{x}'-\bigl(\mathbf{y}(\mathbf{x})-\mathbf{y}(\mathbf{x}')\bigr)\right). \end{equation}
Substituting~(\ref{eqn:moreau_gradient_difference}) into~(\ref{eqn:moreau_hypomonotonicity}) and rearranging gives $2\left\langle\mathbf{x}-\mathbf{x}',\mathbf{y}(\mathbf{x})-\mathbf{y}(\mathbf{x}')\right\rangle\geq\|\mathbf{y}(\mathbf{x})-\mathbf{y}(\mathbf{x}')\|_2^2. $
Consequently,
\begin{equation}\label{eqn:moreau_residual_nonexpansive} 
\begin{split}
    \left\|\mathbf{x}-\mathbf{x}'-\bigl(\mathbf{y}(\mathbf{x})-\mathbf{y}(\mathbf{x}')\bigr)\right\|_2^2
    &=\|\mathbf{x}-\mathbf{x}'\|_2^2-2\left\langle\mathbf{x}-\mathbf{x}',\mathbf{y}(\mathbf{x})-\mathbf{y}(\mathbf{x}')\right\rangle+\|\mathbf{y}(\mathbf{x})-\mathbf{y}(\mathbf{x}')\|_2^2\\
    &\leq\|\mathbf{x}-\mathbf{x}'\|_2^2. 
\end{split}
\end{equation}
Using~(\ref{eqn:moreau_gradient_difference}) and~(\ref{eqn:moreau_residual_nonexpansive}), we obtain
\begin{equation}\label{eqn:moreau_gradient_lipschitz} \|\mathbf{G}(\mathbf{x})-\mathbf{G}(\mathbf{x}')\|_2=2L\left\|\mathbf{x}-\mathbf{x}'-\bigl(\mathbf{y}(\mathbf{x})-\mathbf{y}(\mathbf{x}')\bigr)\right\|_2\leq2L\|\mathbf{x}-\mathbf{x}'\|_2. \end{equation}
Combining~(\ref{eqn:moreau_gradient_formula}), (\ref{eqn:moreau_gradient_norm_bound}), and~(\ref{eqn:moreau_gradient_lipschitz}) proves the result.

\subsection{Proof of Lemma~\ref{lem:perturbed_moreau_descent}}\label{app:proof_perturbed_moreau_descent}

Since $\|\nabla_{\w}\phi(\w,\q)\|_2\leq G$, the function $\phi(\cdot,\q)$ is $G$-Lipschitz for every $\q\in\Delta_m$. As the pointwise maximum of $G$-Lipschitz functions, $\Phi$ is also $G$-Lipschitz. Moreover, Lemma~\ref{lem:Phi_weakly_convex} shows that $\Phi$ is $L$-weakly convex. Hence, Lemma~\ref{lem:moreau_envelope_smooth} applies.

For every $t\in[T]$, define the proximal point
\begin{equation}\label{eqn:generic_moreau_prox_point_new}
\y_t=\argmin_{\y\in\mathbb R^d}\left\{\Phi(\y)+L\|\y-\u_t\|_2^2\right\},
\qquad
\G_t^\Phi:=\nabla\Phi_{1/(2L)}(\u_t)=2L(\u_t-\y_t).
\end{equation}
By the $L$-smoothness of $\phi(\cdot,\q_t)$,
\begin{equation}\label{eqn:generic_moreau_inner_1_new}
\left\langle\nabla_{\w}\phi(\u_t,\q_t),\u_t-\y_t\right\rangle\geq\phi(\u_t,\q_t)-\phi(\y_t,\q_t)-\frac{L}{2}\|\u_t-\y_t\|_2^2.
\end{equation}
By definition, $\phi(\u_t,\q_t)=\Phi(\u_t)-\varepsilon_t^u$, while $\phi(\y_t,\q_t)\leq\Phi(\y_t)$. In addition, the optimality of $\y_t$ in~(\ref{eqn:generic_moreau_prox_point_new}) gives $\Phi(\y_t)+L\|\y_t-\u_t\|_2^2\leq\Phi(\u_t),$ and consequently $\Phi(\u_t)-\Phi(\y_t)\geq L\|\u_t-\y_t\|_2^2.$
Multiplying~(\ref{eqn:generic_moreau_inner_1_new}) by $2L$ and using these relations yields
\begin{equation}\label{eqn:generic_moreau_inner_2_new}
	\begin{split}
    \langle \nabla_{\w}\phi(\u_t,\q_t),\G_t^\Phi\rangle
    &= 2L\langle \nabla_{\w}\phi(\u_t,\q_t),\mathbf{u}_t-\mathbf{y}_t\rangle \\
    &\ge 2L\phi(\mathbf{u}_t,\q_t)-2L\phi(\mathbf{y}_t,\q_t)-L^2\|\mathbf{u}_t-\mathbf{y}_t\|_2^2 \\
    &\ge 2L\phi(\mathbf{u}_t,\q_t)-2L\Phi(\mathbf{y}_t)-L^2\|\mathbf{u}_t-\mathbf{y}_t\|_2^2 \\
    &=2L\Phi(\mathbf{u}_t)-2L\varepsilon_t^u-2L\Phi(\mathbf{y}_t)-L^2\|\mathbf{u}_t-\mathbf{y}_t\|_2^2 \\
    &=
    2L\bigl(\Phi(\mathbf{u}_t)-\Phi(\mathbf{y}_t)\bigr)-L^2\|\mathbf{u}_t-\mathbf{y}_t\|_2^2-2L\varepsilon_t^u \\
    &\ge 2L^2\|\mathbf{u}_t-\mathbf{y}_t\|_2^2-L^2\|\mathbf{u}_t-\mathbf{y}_t\|_2^2-2L\varepsilon_t^u \\
    &=L^2\|\mathbf{u}_t-\mathbf{y}_t\|_2^2-2L\varepsilon_t^u \\
    &=\frac{1}{4}\|\G_t^\Phi\|_2^2-2L\varepsilon_t^u,
	\end{split}
\end{equation}
The $L$-smoothness of $\phi(\cdot,\q_t)$ also gives
\begin{equation*}
\|\nabla_{\w}\phi(\w_t,\q_t)-\nabla_{\w}\phi(\u_t,\q_t)\|_2\leq L\|\w_t-\u_t\|_2=Ld_t.
\end{equation*}
Combining this inequality with~(\ref{eqn:generic_moreau_inner_2_new}) and the bound $\|\G_t^\Phi\|_2\leq G$ from Lemma~\ref{lem:moreau_envelope_smooth}, we obtain
\begin{equation}\label{eqn:generic_moreau_inner_3_new}
\begin{aligned}
\left\langle\nabla_{\w}\phi(\w_t,\q_t),\G_t^\Phi\right\rangle
&=\left\langle\nabla_{\w}\phi(\u_t,\q_t),\G_t^\Phi\right\rangle+\left\langle\nabla_{\w}\phi(\w_t,\q_t)-\nabla_{\w}\phi(\u_t,\q_t),\G_t^\Phi\right\rangle\\
&\geq\frac{1}{4}\|\G_t^\Phi\|_2^2-2L\varepsilon_t^u-Ld_t\|\G_t^\Phi\|_2\\
&\geq\frac{1}{4}\|\G_t^\Phi\|_2^2-2L\varepsilon_t^u-LGd_t.
\end{aligned}
\end{equation}

Using the definition of the Moreau envelope and the update in~(\ref{eqn:generic_virtual_update_new}), we have
\begin{equation}
    \begin{split}
        \Phi_{1/(2L)}(\u_{t+1})
        &= \min_{\y\in\R^d}\{\Phi(\y)+L\|\y-\u_{t+1}\|_2^2\} \\
        &\le \Phi(\y_t)+L\|\y_t-\u_{t+1}\|_2^2 \\
        &= \Phi(\y_t)+L\|\y_t-\u_t+\eta_w\g_{w,t}\|_2^2 \\
        &= \Phi(\y_t)+L\|\y_t-\u_t\|_2^2
            +2L\eta_w\langle \y_t-\u_t,\g_{w,t}\rangle
            +L\eta_w^2\|\g_{w,t}\|_2^2 \\
        &= \Phi_{1/(2L)}(\u_t)
            -\eta_w\langle \g_{w,t},\G_t^\Phi\rangle
            +L\eta_w^2\|\g_{w,t}\|_2^2.
        \label{eqn:generic_moreau_one_step_new}
    \end{split}
\end{equation}
Using $\mathbb E_t[\g_{w,t}]=\nabla_{\w}\phi(\w_t,\q_t)$ and $\mathbb E_t[\|\g_{w,t}\|_2^2]\leq\Gamma_w^2$, taking the conditional expectation of~(\ref{eqn:generic_moreau_one_step_new}) yields
\begin{equation}\label{eqn:moreau_one_step_conditional} 
\begin{split}
    \E_t[\Phi_{1/(2L)}(\u_{t+1})]
    &\leq\Phi_{1/(2L)}(\u_t)-\eta_w\left\langle\nabla_{\w}\phi(\w_t,\q_t),\G_t^\Phi\right\rangle+L\eta_w^2\E_t[\|\g_{w,t}\|_2^2]\\
    &\leq\Phi_{1/(2L)}(\u_t)-\eta_w\left\langle\nabla_{\w}\phi(\w_t,\q_t),\G_t^\Phi\right\rangle+L\eta_w^2\Gamma_w^2 \\
    &\overset{(\ref{eqn:generic_moreau_inner_3_new})}{\leq}\Phi_{1/(2L)}(\u_t)-\frac{\eta_w}{4}\|\G_t^\Phi\|_2^2+2L\eta_w\varepsilon_t^u+LG\eta_w d_t+L\eta_w^2\Gamma_w^2,
\end{split}
\end{equation}
where we use
$$\mathbb E_t[\|\g_{w,t}\|_2^2]
=\left\|\nabla_{\w}\phi(\w_t,\q_t)\right\|_2^2+\mathbb E_t\left[\left\|\g_{w,t}-\nabla_{\w}\phi(\w_t,\q_t)\right\|_2^2\right]\leq G^2+\sigma_w^2=\Gamma_w^2.$$
Taking the total expectation on both sides of~(\ref{eqn:moreau_one_step_conditional}) and rearranging the stationarity term gives
\begin{equation}\label{eqn:generic_moreau_one_step_expectation_new}
\frac{\eta_w}{4}\mathbb E[\|\G_t^\Phi\|_2^2]\leq\mathbb E[\Phi_{1/(2L)}(\u_t)]-\mathbb E[\Phi_{1/(2L)}(\u_{t+1})]+2L\eta_w\mathbb E[\varepsilon_t^u]+LG\eta_w\mathbb E[d_t]+L\eta_w^2\Gamma_w^2.
\end{equation}

We next sum~(\ref{eqn:generic_moreau_one_step_expectation_new}) only over $t\in\mathcal J$. Let $F_t=\Phi_{1/(2L)}(\u_t)$. By Lemma~\ref{lem:moreau_envelope_smooth}, $\|\nabla\Phi_{1/(2L)}(\x)\|_2\leq G$, and hence $\Phi_{1/(2L)}$ is $G$-Lipschitz. Therefore, for every $t\in\mathcal S$,
\begin{equation}\label{eqn:generic_skipped_envelope_increment_new}
\begin{aligned}
\mathbb E[F_{t+1}-F_t]\leq G\mathbb E[\|\u_{t+1}-\u_t\|_2]=G\eta_w\mathbb E[\|\g_{w,t}\|_2]\leq G\eta_w\sqrt{\mathbb E[\|\g_{w,t}\|_2^2]}\leq G\eta_w\Gamma_w.
\end{aligned}
\end{equation}
Since $\sum_{t\in\mathcal J}(F_t-F_{t+1})=F_1-F_{T+1}+\sum_{t\in\mathcal S}(F_{t+1}-F_t)$ and $F_{T+1}\geq\Phi_*$, it follows from~(\ref{eqn:generic_skipped_envelope_increment_new}) that
\begin{equation}\label{eqn:generic_subset_telescope_new}
\sum_{t\in\mathcal J}\left(\mathbb E[F_t]-\mathbb E[F_{t+1}]\right)\leq\Delta_u+G\eta_w\Gamma_w|\mathcal S|.
\end{equation}
Summing~(\ref{eqn:generic_moreau_one_step_expectation_new}) over $t\in\mathcal J$, applying~(\ref{eqn:generic_subset_telescope_new}), and using $|\mathcal J|\leq T$ yields
\begin{equation}\label{eqn:generic_subset_stationarity_new}
\frac{1}{T}\sum_{t\in\mathcal J}\mathbb E[\|\G_t^\Phi\|_2^2]
\leq\frac{4\Delta_u}{\eta_wT}+\frac{8L}{T}\sum_{t\in\mathcal J}\mathbb E[\varepsilon_t^u]+\frac{4LG}{T}\sum_{t\in\mathcal J}\mathbb E[d_t]+4L\eta_w\Gamma_w^2+\frac{4G\Gamma_w|\mathcal S|}{T}.
\end{equation}
Finally, Lemma~\ref{lem:moreau_envelope_smooth} and $G\leq\Gamma_w$ imply $\sum_{t\in\mathcal S}\mathbb E[\|\G_t^\Phi\|_2^2]\leq G^2|\mathcal S|\leq\Gamma_w^2|\mathcal S|.$ Adding this bound to~(\ref{eqn:generic_subset_stationarity_new}) and using $4G\Gamma_w+\Gamma_w^2\leq5\Gamma_w^2$ proves~(\ref{eqn:generic_moreau_descent_new}).

\subsection{Proof of Lemma~\ref{lem:blockwise_dual_gap}}\label{app:proof_blockwise_dual_gap}
Let $\q^{\mathrm{unif}}=m^{-1}\mathbf1_m$. For every block $\mathcal I_c$, choose an anchor $\q_{s_c}^*\in\argmax_{\q\in\Delta_m}\phi(\w_{s_c},\q).$
For every $t\in\mathcal I_c$, decompose the instantaneous dual gap as
\begin{equation}\label{eqn:blockwise_gap_decomposition_new}
\varepsilon_t^q=\underbrace{\Phi(\w_t)-\phi(\w_t,\q_{s_c}^*)}_{A_t}+\underbrace{\phi(\w_t,\q_{s_c}^*)-\phi(\w_t,\q_t)}_{B_t}.
\end{equation}

We first control the error induced by fixing the anchor within each block. Let $\q_t^*\in\argmax_{\q\in\Delta_m}\phi(\w_t,\q)$. By the optimality of $\q_{s_c}^*$ at $\w_{s_c}$, we obtain $\phi(\w_{s_c},\q_t^*)\leq\phi(\w_{s_c},\q_{s_c}^*).$
Since every $R_i$ is $G$-Lipschitz, $\phi(\cdot,\q)$ is $G$-Lipschitz uniformly over $\q\in\Delta_m$. Hence,
\begin{equation*}\label{eqn:blockwise_anchor_movement_new}
\begin{aligned}
A_t
&=\phi(\w_t,\q_t^*)-\phi(\w_t,\q_{s_c}^*)\\
&=\phi(\w_t,\q_t^*)-\phi(\w_{s_c},\q_t^*)+\phi(\w_{s_c},\q_t^*)-\phi(\w_{s_c},\q_{s_c}^*)+\phi(\w_{s_c},\q_{s_c}^*)-\phi(\w_t,\q_{s_c}^*)\\
&\leq\phi(\w_t,\q_t^*)-\phi(\w_{s_c},\q_t^*)+\phi(\w_{s_c},\q_{s_c}^*)-\phi(\w_t,\q_{s_c}^*)\\
&\leq2G\|\w_t-\w_{s_c}\|_2.
\end{aligned}
\end{equation*}
For $t\in\mathcal I_c$, the triangle inequality gives $\|\w_t-\w_{s_c}\|_2\leq\sum_{\tau=s_c}^{t-1}\|\w_{\tau+1}-\w_\tau\|_2.$
Since $|\mathcal I_c|\leq B$, summing over each block yields
\begin{equation}\label{eqn:blockwise_anchor_sum_new}
\sum_{c=1}^N\sum_{t\in\I_c}\E[A_t]\leq2GB\sum_{c=1}^N\sum_{\tau=s_c}^{e_c-1}\E[\|\w_{\tau+1}-\w_\tau\|_2]\leq2GB\sum_{t=1}^{T-1}\E[\|\w_{t+1}-\w_t\|_2].
\end{equation}

We next control $B_t$ using KL-based online mirror ascent. Let $\boldsymbol\xi_t=\widehat{\mathbf R}_t-\mathbf R(\w_t)$, where $\xi_{t,i}=\ell(\w_t;\z_t^{(i)})-R_i(\w_t)$. Since $\w_t$ is $\mathcal F_t$-measurable, Assumption~\ref{ass:nonconvex_dual_oracle} implies
\begin{equation}\label{eqn:blockwise_conditional_dual_oracle_new}
\mathbb E_t[\xi_{t,i}]=0,\qquad \mathbb E_t\left[\exp(\lambda\xi_{t,i})\right]\leq\exp\left(\frac{\lambda^2\sigma_q^2}{2}\right),\qquad\forall i\in[m],\ \lambda\in\mathbb R.
\end{equation}
The first-order optimality condition for~(\ref{eqn:blockwise_mirror_iterate_new}), together with the Bregman three-point identity, implies that for every $\q\in\Delta_m$,
\begin{equation}\label{eqn:blockwise_three_point_new}
\eta_q\langle\widehat{\mathbf R}_t,\q-\bar{\q}_{t+1}\rangle\leq D_{\mathrm{KL}}(\q\|\q_t)-D_{\mathrm{KL}}(\q\|\bar{\q}_{t+1})-D_{\mathrm{KL}}(\bar{\q}_{t+1}\|\q_t).
\end{equation}

Since $\phi$ is linear in $\q$, and both $\q_{s_c}^*$ and $\q_t$ are $\mathcal F_t$-measurable, the inequality~(\ref{eqn:blockwise_three_point_new}) with $\q=\q_{s_c}^*$ yields
\begin{equation}\label{eqn:blockwise_B_one_step_pre_noise_new}
\begin{aligned}
B_t
&=\left\langle\mathbf R(\w_t),\q_{s_c}^*-\q_t\right\rangle\\
&=\mathbb E_t\left[\left\langle\widehat{\mathbf R}_t,\q_{s_c}^*-\bar{\q}_{t+1}\right\rangle+\left\langle\widehat{\mathbf R}_t,\bar{\q}_{t+1}-\q_t\right\rangle\right]\\
&\leq\frac{D_{\mathrm{KL}}(\q_{s_c}^*\|\q_t)-\mathbb E_t[D_{\mathrm{KL}}(\q_{s_c}^*\|\bar{\q}_{t+1})]}{\eta_q}-\frac{1}{\eta_q}\mathbb E_t[D_{\mathrm{KL}}(\bar{\q}_{t+1}\|\q_t)]\\
&\quad+\mathbb E_t\left[\left\langle\mathbf R(\w_t),\bar{\q}_{t+1}-\q_t\right\rangle+\left\langle\boldsymbol\xi_t,\bar{\q}_{t+1}-\q_t\right\rangle\right]\\
&=\frac{D_{\mathrm{KL}}(\q_{s_c}^*\|\q_t)-\mathbb E_t[D_{\mathrm{KL}}(\q_{s_c}^*\|\bar{\q}_{t+1})]}{\eta_q}+\mathbb E_t\left[\phi(\w_t,\bar{\q}_{t+1})-\phi(\w_t,\q_t)\right]\\
&\quad+\mathbb E_t\left[\left\langle\boldsymbol\xi_t,\bar{\q}_{t+1}-\q_t\right\rangle-\frac{1}{\eta_q}D_{\mathrm{KL}}(\bar{\q}_{t+1}\|\q_t)\right].
\end{aligned}
\end{equation}

We next control the final term in~(\ref{eqn:blockwise_B_one_step_pre_noise_new}) through the conjugacy between negative entropy and log-sum-exp. 
Since the mirror-ascent iterates remain in the relative interior of $\Delta_m$, for every $\boldsymbol\xi\in\mathbb R^m$,
\begin{equation}\label{eqn:blockwise_entropy_conjugate_new}
\sup_{\q\in\Delta_m}\left\{\langle\boldsymbol\xi,\q-\q_t\rangle-\frac{1}{\eta_q}D_{\mathrm{KL}}(\q\|\q_t)\right\}=\frac{1}{\eta_q}\ln\left(\sum_{i=1}^m q_{t,i}\exp(\eta_q\xi_i)\right)-\langle\boldsymbol\xi,\q_t\rangle.
\end{equation}
Since $\bar{\q}_{t+1}\in\Delta_m$, applying~(\ref{eqn:blockwise_entropy_conjugate_new}) with $\boldsymbol\xi=\boldsymbol\xi_t$ and taking the conditional expectation gives
\begin{equation}\label{eqn:blockwise_noise_control_new}
\begin{aligned}
&\mathbb E_t\left[\left\langle\boldsymbol\xi_t,\bar{\q}_{t+1}-\q_t\right\rangle-\frac{1}{\eta_q}D_{\mathrm{KL}}(\bar{\q}_{t+1}\|\q_t)\right]\\
\leq{}&\frac{1}{\eta_q}\mathbb E_t\left[\ln\left(\sum_{i=1}^m q_{t,i}\exp(\eta_q\xi_{t,i})\right)\right]-\mathbb E_t[\langle\boldsymbol\xi_t,\q_t\rangle]\\
\leq{}&\frac{1}{\eta_q}\ln\left(\sum_{i=1}^m q_{t,i}\mathbb E_t[\exp(\eta_q\xi_{t,i})]\right)\leq\frac{\eta_q\sigma_q^2}{2}.
\end{aligned}
\end{equation}
The second inequality uses conditional Jensen's inequality, the $\mathcal F_t$-measurability of $\q_t$, and $\mathbb E_t[\boldsymbol\xi_t]=\mathbf0_m$. The last inequality follows from~(\ref{eqn:blockwise_conditional_dual_oracle_new}) and $\sum_{i=1}^m q_{t,i}=1$. Combining~(\ref{eqn:blockwise_B_one_step_pre_noise_new}) and~(\ref{eqn:blockwise_noise_control_new}) gives
\begin{equation}\label{eqn:blockwise_B_one_step_new}
B_t
\leq\frac{D_{\mathrm{KL}}(\q_{s_c}^*\|\q_t)-\mathbb E_t[D_{\mathrm{KL}}(\q_{s_c}^*\|\bar{\q}_{t+1})]}{\eta_q}+\mathbb E_t\left[\phi(\w_t,\bar{\q}_{t+1})-\phi(\w_t,\q_t)\right]+\frac{\eta_q\sigma_q^2}{2}.
\end{equation}
Taking the total expectation and summing~(\ref{eqn:blockwise_B_one_step_new}) over $t\in\mathcal I_c$, the KL-divergence terms telescope because $\bar{\q}_{t+1}=\q_{t+1}$ for every $t<e_c$. Dropping the final nonnegative divergence gives
\begin{equation}\label{eqn:blockwise_B_block_new}
\sum_{t\in\mathcal I_c}\mathbb E[B_t]\leq\frac{\mathbb E[D_{\mathrm{KL}}(\q_{s_c}^*\|\q_{s_c})]}{\eta_q}+\mathbb E[J_c]+\frac{\eta_q\sigma_q^2}{2}|\mathcal I_c|,
\end{equation}
where $J_c:=\sum_{t=s_c}^{e_c}\left(\phi(\w_t,\bar{\q}_{t+1})-\phi(\w_t,\q_t)\right).$ Since $\q_{s_c}=m^{-1}\1_m$, we obtain
\begin{equation*}
D_{\mathrm{KL}}(\q_{s_c}^*\|\q_{s_c})=\sum_{i=1}^m q_{s_c,i}^*\ln(mq_{s_c,i}^*)=\ln m+\sum_{i=1}^m q_{s_c,i}^*\ln q_{s_c,i}^*\leq\ln m.
\end{equation*}
It remains to control the increments $J_c$. Since $\bar{\q}_{t+1}=\q_{t+1}$ for every $t<e_c$ and $\q_{s_c}=\q^{\mathrm{unif}}$, rearranging the sum gives
\begin{equation}\label{eqn:blockwise_J_telescope_new}
J_c=\phi(\w_{e_c},\bar{\q}_{e_c+1})-\phi(\w_{s_c},\q^{\mathrm{unif}})+\sum_{t=s_c}^{e_c-1}\left(\phi(\w_t,\q_{t+1})-\phi(\w_{t+1},\q_{t+1})\right).
\end{equation}
For every nonfinal block $c<N$, we have $s_{c+1}=e_c+1$ and $\q_{s_{c+1}}=\q^{\mathrm{unif}}$. Since $\phi(\w_{e_c},\bar{\q}_{e_c+1})\leq\Phi(\w_{e_c})$ and both $\Phi$ and $\phi(\cdot,\q)$ are $G$-Lipschitz,
\begin{equation}\label{eqn:blockwise_J_nonfinal_derivation_new}
\begin{aligned}
J_c
&\leq\Phi(\w_{e_c})-\phi(\w_{s_c},\q^{\mathrm{unif}})+G\sum_{t=s_c}^{e_c-1}\|\w_{t+1}-\w_t\|_2\\
&\leq\Phi(\w_{e_c+1})-\phi(\w_{e_c+1},\q^{\mathrm{unif}})+G\|\w_{e_c+1}-\w_{e_c}\|_2+G\|\w_{e_c+1}-\w_{s_c}\|_2\\
&\quad+G\sum_{t=s_c}^{e_c-1}\|\w_{t+1}-\w_t\|_2\\
&\leq\varepsilon_{s_{c+1}}^q+2G\sum_{t=s_c}^{e_c}\|\w_{t+1}-\w_t\|_2.
\end{aligned}
\end{equation}
For the final block, $e_N=T$. Applying~(\ref{eqn:blockwise_J_telescope_new}), $\phi(\w_T,\bar{\q}_{T+1})\leq\Phi(\w_T)$, and the same Lipschitz bounds gives
\begin{equation}\label{eqn:blockwise_J_final_new}
\begin{aligned}
J_N
&\leq\Phi(\w_T)-\phi(\w_{s_N},\q^{\mathrm{unif}})+G\sum_{t=s_N}^{T-1}\|\w_{t+1}-\w_t\|_2\\
&\leq\Phi(\w_1)-\phi(\w_1,\q^{\mathrm{unif}})+G\|\w_T-\w_1\|_2+G\|\w_{s_N}-\w_1\|_2+G\sum_{t=s_N}^{T-1}\|\w_{t+1}-\w_t\|_2\\
&\leq \varepsilon_1^q+2G\sum_{t=1}^{T-1}\|\w_{t+1}-\w_t\|_2.
\end{aligned}
\end{equation}
Summing~(\ref{eqn:blockwise_J_nonfinal_derivation_new}) over $c<N$, adding~(\ref{eqn:blockwise_J_final_new}), and taking expectations yields
\begin{equation}\label{eqn:blockwise_J_sum_new}
\sum_{c=1}^N\mathbb E[J_c]\leq\sum_{t\in\mathcal S}\mathbb E[\varepsilon_t^q]+4G\sum_{t=1}^{T-1}\mathbb E[\|\w_{t+1}-\w_t\|_2].
\end{equation}

Combining~(\ref{eqn:blockwise_gap_decomposition_new}), (\ref{eqn:blockwise_anchor_sum_new}), (\ref{eqn:blockwise_B_block_new}), and~(\ref{eqn:blockwise_J_sum_new}), and using $\sum_{c=1}^N|\mathcal I_c|=T$, gives
\begin{equation*}\label{eqn:blockwise_before_cancel_new}
\sum_{t=1}^T\mathbb E[\varepsilon_t^q]\leq2G(B+2)\sum_{t=1}^{T-1}\mathbb E[\|\w_{t+1}-\w_t\|_2]+\frac{N\ln m}{\eta_q}+\frac{\eta_q\sigma_q^2T}{2}+\sum_{t\in\mathcal S}\mathbb E[\varepsilon_t^q].
\end{equation*}
Subtracting the block-start gaps from both sides gives
\begin{equation*}
\sum_{t\in\mathcal J}\mathbb E[\varepsilon_t^q]\leq2G(B+2)\sum_{t=1}^{T-1}\mathbb E[\|\w_{t+1}-\w_t\|_2]+\frac{N\ln m}{\eta_q}+\frac{\eta_q\sigma_q^2T}{2}.
\end{equation*}
Finally, $N=\lceil T/B\rceil\leq T/B+1\leq2T/B$ because $B\leq T$. Dividing the preceding inequality by $T$ proves~(\ref{eqn:blockwise_dual_gap_general_new}).

\subsection{Proof of Lemma~\ref{lem:compression_residuals}}\label{app:proof_compression_residuals}
Let $\rho=\sqrt{1-\delta}\in[0,1)$. By the definition of a $\delta$-approximate compressor, for every deterministic $\x\in\mathbb R^d$,
\begin{equation}\label{eqn:compression_first_moment}
\mathbb E_{\mathcal C}\left[\|\mathcal C(\x)-\x\|_2\right]\leq\sqrt{\mathbb E_{\mathcal C}\left[\|\mathcal C(\x)-\x\|_2^2\right]}\leq\rho\|\x\|_2.
\end{equation}
Moreover, the same inequality holds conditionally for any random input independent of the fresh compressor realization. 

\paragraph{General compressors.} 
We first control the aggregated uplink residual. Define $S_t=\sum_{i=1}^m\mathbb E[\|\e_t^{(i)}\|_2]$. The uplink error-feedback recursion in Algorithm~\ref{alg:aflcom_br} and~(\ref{eqn:compression_first_moment}) give
\begin{equation}\label{eqn:compression_uplink_recursion}
\begin{aligned}
\mathbb E\left[\|\e_{t+1}^{(i)}\|_2\right]
&=\mathbb E\left[\left\|\e_t^{(i)}+q_{t,i}\g_{w,t}^{(i)}-\mathcal C\left(\e_t^{(i)}+q_{t,i}\g_{w,t}^{(i)}\right)\right\|_2\right]\\
&\leq\rho\mathbb E\left[\left\|\e_t^{(i)}+q_{t,i}\g_{w,t}^{(i)}\right\|_2\right]\\
&\leq\rho\mathbb E\left[\|\e_t^{(i)}\|_2\right]+\rho\mathbb E\left[q_{t,i}\|\g_{w,t}^{(i)}\|_2\right].
\end{aligned}
\end{equation}
Assumption~\ref{ass:nonconvex_primal_oracle} and conditional Jensen's inequality imply $\mathbb E_t[\|\g_{w,t}^{(i)}\|_2]\leq\Gamma_w$. 
Summing~(\ref{eqn:compression_uplink_recursion}) over $i\in[m]$ and applying $\sum_{i=1}^m\mathbb E[q_{t,i}\|\g_{w,t}^{(i)}\|_2]\leq\Gamma_w$ yields
\begin{equation}\label{eqn:compression_uplink_sum_recursion}
S_{t+1}\leq\rho S_t+\rho\Gamma_w.
\end{equation}
Because $\e_1^{(i)}=\mathbf0_d$ for all $i\in[m]$, iterating~(\ref{eqn:compression_uplink_sum_recursion}) gives
\begin{equation*}\label{eqn:compression_uplink_sum_bound}
S_{t+1}\leq\sum_{k=1}^t\rho^k\Gamma_w\leq\frac{\rho}{1-\rho}\Gamma_w,\qquad t\leq T.
\end{equation*}
Recall that $\e_t=\sum_{i=1}^m\e_t^{(i)}$. By the triangle inequality,
\begin{equation}\label{eqn:compression_uplink_residual_bound}
\mathbb E\left[\|\e_t\|_2\right]\leq S_t\leq\frac{\rho}{1-\rho}\Gamma_w\leq\frac{2-\delta}{\delta}\Gamma_w,\qquad t\leq T+1,
\end{equation}
where the last inequality follows from
\begin{equation*}
\frac{\rho}{1-\rho}=\frac{\rho(1+\rho)}{\delta}=\frac{\sqrt{1-\delta}+1-\delta}{\delta}\leq\frac{2-\delta}{\delta}.
\end{equation*}

We next control the downlink residual. Define the uncompressed weighted stochastic gradient $\g_{w,t}=\sum_{i=1}^m q_{t,i}\g_{w,t}^{(i)}.$
Summing the uplink error-feedback recursions over the workers gives
\begin{equation*}\label{eqn:compression_uplink_message_identity}
\hatgb_{w,t}=\sum_{i=1}^m\Delta_{w,t}^{(i)}=\g_{w,t}+\e_t-\e_{t+1}.
\end{equation*}
Combining $\mathbb E[\|\g_{w,t}\|_2]\leq\Gamma_w$ with~(\ref{eqn:compression_uplink_residual_bound}) yields
\begin{equation}\label{eqn:compression_uplink_message_bound}
\begin{aligned}
\mathbb E\left[\|\hatgb_{w,t}\|_2\right]
&\leq\mathbb E\left[\|\g_{w,t}\|_2\right]+\mathbb E\left[\|\e_t\|_2\right]+\mathbb E\left[\|\e_{t+1}\|_2\right]\\
&\leq\left(1+\frac{2\rho}{1-\rho}\right)\Gamma_w
=\frac{1+\rho}{1-\rho}\Gamma_w.
\end{aligned}
\end{equation}
The downlink error-feedback recursion and~(\ref{eqn:compression_first_moment}) imply
\begin{equation}\label{eqn:compression_downlink_recursion}
\begin{aligned}
\mathbb E\left[\|\widehat\e_{t+1}\|_2\right]
&=\mathbb E\left[\left\|\hateb_t+\hatgb_{w,t}-\mathcal C\left(\hateb_t+\hatgb_{w,t}\right)\right\|_2\right]\\
&\leq\rho\mathbb E\left[\|\hateb_t+\hatgb_{w,t}\|_2\right]\\
&\leq\rho\mathbb E\left[\|\hateb_t\|_2\right]+\rho\mathbb E\left[\|\hatgb_{w,t}\|_2\right].
\end{aligned}
\end{equation}
Using $\widehat\e_1=\mathbf0_d$, substituting~(\ref{eqn:compression_uplink_message_bound}), and iterating~(\ref{eqn:compression_downlink_recursion}), we obtain
\begin{equation*}
\begin{aligned}
\mathbb E\left[\|\hateb_t\|_2\right]\leq\frac{\rho}{1-\rho}\frac{1+\rho}{1-\rho}\Gamma_w=\frac{\rho(1+\rho)}{(1-\rho)^2}\Gamma_w=\frac{\rho(1+\rho)^3}{\delta^2}\Gamma_w\leq\frac{8}{\delta^2}\Gamma_w,\qquad t\leq T+1.
\end{aligned}
\end{equation*}
Since $\rb_t=\e_t+\hateb_t$, the two uniform residual bounds imply
\begin{equation*}
\frac{1}{T}\sum_{t=1}^T\mathbb E\left[\|\rb_t\|_2\right]\leq\frac{2-\delta}{\delta}\Gamma_w+\frac{8}{\delta^2}\Gamma_w=\frac{8+2\delta-\delta^2}{\delta^2}\Gamma_w.
\end{equation*}
Moreover, the error-feedback identity $\Delta_{w,t}=\g_{w,t}+\rb_t-\rb_{t+1}$ and $\mathbb E[\|\g_{w,t}\|_2]\leq\Gamma_w$ give
\begin{equation*}
\frac{1}{T}\sum_{t=1}^T\mathbb E\left[\|\Delta_{w,t}\|_2\right]\leq\Gamma_w+2\left(\frac{2-\delta}{\delta}+\frac{8}{\delta^2}\right)\Gamma_w=\frac{16+4\delta-\delta^2}{\delta^2}\Gamma_w.
\end{equation*}

\paragraph{Additive-and-idempotent compressors with shared randomness.} 
Suppose that $\mathcal C\in\mathcal C_{\mathrm{AI}}$ and that the server and workers use the same compressor realization at round $t$. Denote this realization by $\mathcal C^{(\xi_t)}$. By additivity,
\begin{equation}\label{eqn:compression_ai_uplink_aggregate}
\hatgb_{w,t}
=\sum_{i=1}^m\mathcal C^{(\xi_t)}\left(\e_t^{(i)}+q_{t,i}\g_{w,t}^{(i)}\right)
=\mathcal C^{(\xi_t)}\left(\sum_{i=1}^m\e_t^{(i)}+\sum_{i=1}^m q_{t,i}\g_{w,t}^{(i)}\right).
\end{equation}
Applying idempotence to~(\ref{eqn:compression_ai_uplink_aggregate}) gives the pathwise identity
\begin{equation}\label{eqn:compression_ai_fixed_point}
\mathcal C^{(\xi_t)}\left(\hatgb_{w,t}\right)=\hatgb_{w,t}.
\end{equation}
We now prove by induction that $\hateb_t=\mathbf0_d$ for every $t\leq T+1$. The claim holds at $t=1$ by initialization. Suppose that $\hateb_t=\mathbf0_d$. Using~(\ref{eqn:compression_ai_fixed_point}),
\begin{equation*}
\Delta_{w,t}=\mathcal C^{(\xi_t)}\left(\hateb_t+\hatgb_{w,t}\right)=\mathcal C^{(\xi_t)}\left(\hatgb_{w,t}\right)=\hatgb_{w,t}.
\end{equation*}
Consequently, $\widehat\e_{t+1}=\hateb_t+\hatgb_{w,t}-\Delta_{w,t}=\mathbf0_d.$
Thus, we obtain $\hateb_t=\mathbf0_d$ for all $t\leq T+1$.

Since $\hateb_t=\mathbf0_d$ pathwise, we have $\rb_t=\e_t$ and $\Delta_{w,t}=\g_{w,t}+\e_t-\e_{t+1}$. Therefore,
\begin{equation*}
\frac{1}{T}\sum_{t=1}^T\mathbb E\left[\|\rb_t\|_2\right]\leq\frac{2-\delta}{\delta}\Gamma_w\leq\frac{2\Gamma_w}{\delta},
\end{equation*}
and
\begin{equation}
\frac{1}{T}\sum_{t=1}^T\mathbb E\left[\|\Delta_{w,t}\|_2\right]\leq\Gamma_w+\frac{2(2-\delta)}{\delta}\Gamma_w=\frac{4-\delta}{\delta}\Gamma_w\leq\frac{4\Gamma_w}{\delta}.
\end{equation}

\end{document}